\documentclass{article}

\usepackage{arxiv}

\usepackage[utf8]{inputenc}
\usepackage[T1]{fontenc}    
\usepackage{hyperref}       
\usepackage{url}            
\usepackage{booktabs}       
\usepackage{amsfonts}       
\usepackage{amsmath}
\usepackage{nicefrac}       
\usepackage{microtype}      
\usepackage{cleveref}       
\usepackage{lipsum}         
\usepackage{graphicx}
\usepackage{doi}
\usepackage{caption}
\usepackage{subcaption}
\usepackage{xcolor}

\makeatletter
\def\blfootnote{\xdef\@thefnmark{}\@footnotetext}
\makeatother

\begin{document}

\title{Geometry-Aware Physics-Informed PointNets for Modeling Flows Across Porous Structures}

\author{\href{https://orcid.org/0009-0003-1381-9605}{\includegraphics[scale=0.06]{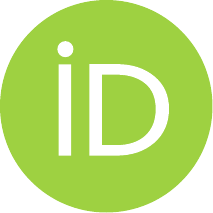}\hspace{1mm}
  Luigi Ciceri}\\
  Universit\`a degli Studi di Milano-Bicocca \\
  Milan, Italy \\
  \texttt{l.ciceri7@campus.unimib.it}\\
  \and
  \href{https://orcid.org/0000-0002-1087-4866}{\includegraphics[scale=0.06]{orcid.pdf}\hspace{1mm}
  Corrado Mio}\\
  Khalifa University of Science and Technology\\
  Abu Dhabi, UAE\\
  \texttt{corrado.mio@ku.ac.ae}\\
  \and
  \href{https://orcid.org/0000-0002-3299-448X}{\includegraphics[scale=0.06]{orcid.pdf}\hspace{1mm}
  Jianyi Lin}\\
  Università Cattolica del Sacro Cuore\\
  Milan, Italy\\
  \texttt{jianyi.lin@unicatt.it}\\
  \and
  \href{https://orcid.org/0000-0001-5186-0199}{\includegraphics[scale=0.06]{orcid.pdf}\hspace{1mm}
  Gabriele Gianini}\\
  Universit\`a degli Studi di Milano-Bicocca \\
  Milan, Italy \\
  \texttt{gabriele.gianini@unimib.it}\\
}

\maketitle
\begin{abstract}
  Predicting flows that occur both through and around porous bodies is challenging due to coupled physics across fluid and porous regions and the need to generalize across diverse geometries and boundary conditions. We address this problem using two Physics Informed learning approaches: Physics Informed PointNets (PIPN) and Physics Informed Geometry Aware Neural Operator (P-IGANO). We enforce the incompressible Navier Stokes equations in the free-flow region and a Darcy Forchheimer extension in the porous region within a unified loss and condition the networks on geometry and material parameters. Datasets are generated with OpenFOAM on 2D ducts containing porous obstacles and on 3D windbreak scenarios with tree canopies and buildings. We first verify the pipeline via the method of manufactured solutions, then assess generalization to unseen shapes, and for PI-GANO, to variable boundary conditions and parameter settings. The results show consistently low velocity and pressure errors in both seen and unseen cases, with accurate reproduction of the wake structures. Performance degrades primarily near sharp interfaces and in regions with large gradients. Overall, the study provides a first systematic evaluation of PIPN/PI-GANO for simultaneous through-and-around porous flows and shows their potential to accelerate design studies without retraining per geometry.
\end{abstract}


\section{Introduction}\label{sec:intro}

\blfootnote{This is the full preprint version of the article accepted by MEDES 2025 as a short paper.}Flows that occur through and around porous bodies at different scales are central to applications such as submerged breakwaters, rock-filled gabions, industrial filters, catalytic beds, heat exchangers, and even biological systems such as coral reefs or plant canopies. Another notable application are windbreaks \cite{mahgoub2021numerical}, such as vegetation barriers, fences, and perforated facades of buildings, used in various environmental and wind mitigation applications.
In these cases, accurate prediction is essential because the interplay between the internal flow through the porous structure and the external flow around it controls transport efficiency, drag forces, and overall system performance.

Modeling such scenarios is doubly complex: the flow through the porous matrix depends on microstructural properties like porosity and permeability, while the flow around the body is influenced by external geometry and boundary layer dynamics. The problem is normally addressed using Computational Fluid Dynamics (CFD) \cite{caucao2020conforming,cimolin2013navier,dalabaev2024currents,dehghan2024fully,parasyris2019computational,rybak2021validation,zampogna2016fluid}, by solving the governing Navier-Stokes fluid equations \cite{wendt2008cfd} (and their extensions) to different regions of a discretized spatial domain (e.g., the Darcy-Forchheimer model for the porous region \cite{dupre2020relating}). Capturing both regimes simultaneously in CFD requires fine spatial resolution and coupled modeling, leading to a high computational cost.
An additional challenge is given by the need to generalize to different geometries: this is necessary for design and optimization workflows, where one must rapidly evaluate many shape variants without rerunning full-scale simulations for each case.

The first two challenges can be addressed using machine learning approaches recently introduced for fluid flow prediction, which, however, have not yet been applied to the simultaneous modeling of flows both through and around porous bodies. Physics-Informed Neural Networks (PINNs) \cite{raissi2019physics} embed the governing equations of fluid flow directly into the training loss, allowing models to learn solutions that satisfy physical laws even when data are sparse.
To address the geometry generalization challenge one can resort to a kind of "geometry-informed" networks, the
PointNet \cite{qi2017pointnet}: these are neural architectures designed to process unstructured point cloud data, which makes them well suited for representing complex geometries without requiring a regular grid.

Physics-Informed PointNets \cite{kashefi2022physics} (PIPN) combine the capabilities of PINNs and PointNets. Furthermore, based on PIPN, the Physics-Informed Geometry-Aware Neural Operator (PI-GANO) \cite{zhong2025pigano} extends the paradigm by learning mappings between geometric configurations and solution fields \textit{in operator form}, allowing for even greater generalization across boundary conditions. The PI-GANO approach, by further adapting the model to incorporate the parameters of the equations, can also predict solutions for previously unseen parameter values, opening opportunities for engineering applications.

A first contribution of the present study consists in addressing, for the first time, the challenge of predicting steady incompressible flows both through and around porous bodies of arbitrary geometries based on the the Navier–Stokes and Darcy-Forchheimer equations by leveraging the PIPN and  PI-GANO architectures, i.e. without retraining for each geometry.

The second contribution consists in extending both the PI-GANO architecture to handle 3D cases which are notoriously expensive to simulate using classic solvers. Moreover the conducted experiments included extremely complex and irregular geometries, such as trees and houses, allowing to test the true capabilities of the architecture in almost real world scenarios.

The experimentation is conducted on CFD-simulated datasets generated with OpenFOAM \cite{jasak2009openfoam}, an open-source computational fluid dynamics toolbox, using stylized porous body and shape configurations. The emphasis of this work is on addressing the computational challenges inherent in this double-flow regime and assessing the performance of different architectures, rather than on the detailed physical modeling of the porous microstructure.

The remainder of this paper is organized as follows.
Section~\ref{sec:equations} introduces the governing equations for steady incompressible flow in porous media and describes the range of geometries considered. The simulation setup, data set generation, and point-cloud conversion are then presented. Section~\ref{sec:PIPN} details the Physics-Informed PointNet and PI-GANO methodologies, including the formulation of the physics-informed loss and the network architectures. Section~\ref{sec:results} reports the results for 2D and 3D porous flow predictions, and for tests of generalization to unseen geometries. Finally, Section~\ref{sec:concolusion} summarizes the main findings and discusses directions for future work.

The source code \cite{ciceri2026} is publicly available.


\section{Fluid Motion Equations and Problem Setup}\label{sec:equations}

We consider a domain consisting of a porous region $\Omega_p$ embedded in a free-fluid region $\Omega_f$, with interface $\partial\Omega_p$ and outer boundary $\partial\Omega_f$ (cf.~\cite{cimolin2013navier}). The governing PDEs can be expressed as $\mathcal{N}_{\Omega_p}[u(x), \lambda, \phi(x)] = 0$, $x \in \Omega_p$, and $\mathcal{N}_{\Omega_f}[u(x), \lambda] = 0$, $x \in \Omega_f$, subject to boundary conditions $\mathcal{B}_{\partial\Omega_p}[u(x)] = g(x)$ and $\mathcal{B}_{\partial\Omega_f}[u(x)] = h(x)$. On $\partial\Omega_p$, velocity and stress continuity are imposed; on $\partial\Omega_f$, standard inflow, outflow and wall conditions are applied. Our goal is to find $u(x)$ satisfying both PDEs, while parameterizing the neural models with respect to geometry $\partial\Omega_p$ and porosity $\phi(x)$ to test generalization to unseen systems.

\subsection{Navier--Stokes and Darcy--Forchheimer Models}
\noindent Consider a fluid described by the incompressible, steady Navier--Stokes equations:
\begin{equation} \label{eq:cont-steady}
  \nabla \cdot \mathbf{u} = 0
\end{equation}
\begin{equation} \label{eq:mom-steady}
  \rho (\mathbf{u}\cdot \nabla)\mathbf{u} = - \nabla p + \mu \nabla^2 \mathbf{u},
\end{equation}
valid in the free-fluid region $\Omega_f$. Here $\mathbf{u}$ is the velocity, $p$ the pressure, $\rho$ the density and $\mu$ the dynamic viscosity.

In the porous region $\Omega_p$, additional drag terms are introduced to represent resistance of the solid matrix. The Darcy--Forchheimer form of Eq. \ref{eq:mom-steady} reads
\begin{equation} \label{eq:forchheimer}
  \rho\, (\mathbf{u}\cdot\nabla)\mathbf{u} = - \nabla p + \mu \nabla^2 \mathbf{u} - \Bigl( \mu D + \tfrac{1}{2}\rho F \lvert \mathbf{u}\rvert \Bigr)\mathbf{u},
\end{equation}
with $D$ the Darcy and $F$ the Forchheimer coefficients. The linear term $\mu D \mathbf{u}$ models viscous drag, while the quadratic term accounts for inertial corrections at moderate pore scale Reynolds numbers~\cite{das2017equations}. For packed spheres these coefficients can be expressed from geometry as a function of the porosity fraction $\phi$
\begin{equation}\label{eq:DandF}
  D = \frac{180}{d_p^2}\frac{(1-\phi)^2}{\phi^3},
  \qquad
  F = \frac{1.8}{d_p}\frac{1-\phi}{\phi^3},
\end{equation}
where $d_p$ is the particle (or Sauter mean) diameter~\cite{kozeny1927,carman1937,carman1956,kaviany2012principles,vafai2015handbook}.

\subsection{2D study cases}
\begin{figure}[tb]
  \centering
  \begin{subfigure}[b]{0.2\textwidth}
    \includegraphics[width=\textwidth]{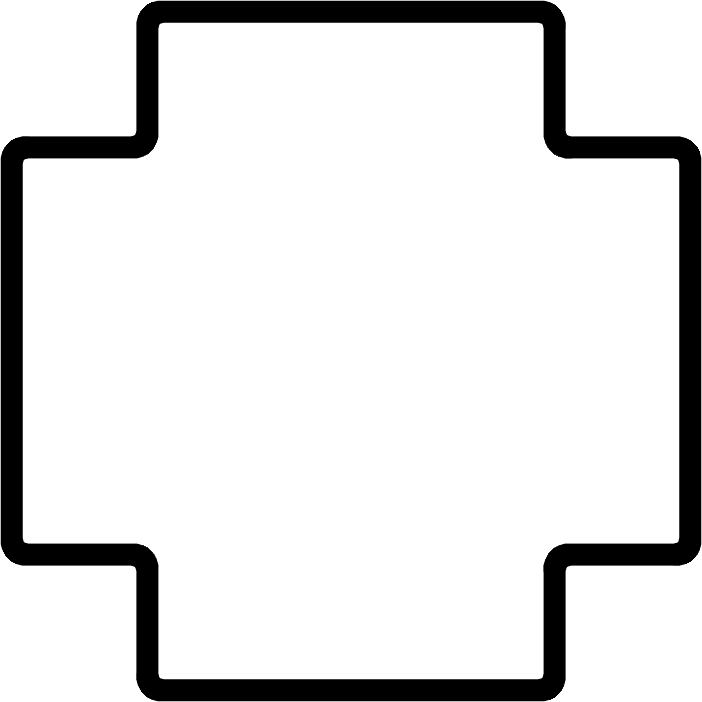}
    \caption{}
  \end{subfigure}
  \hfill
  \begin{subfigure}[b]{0.2\textwidth}
    \includegraphics[width=\textwidth]{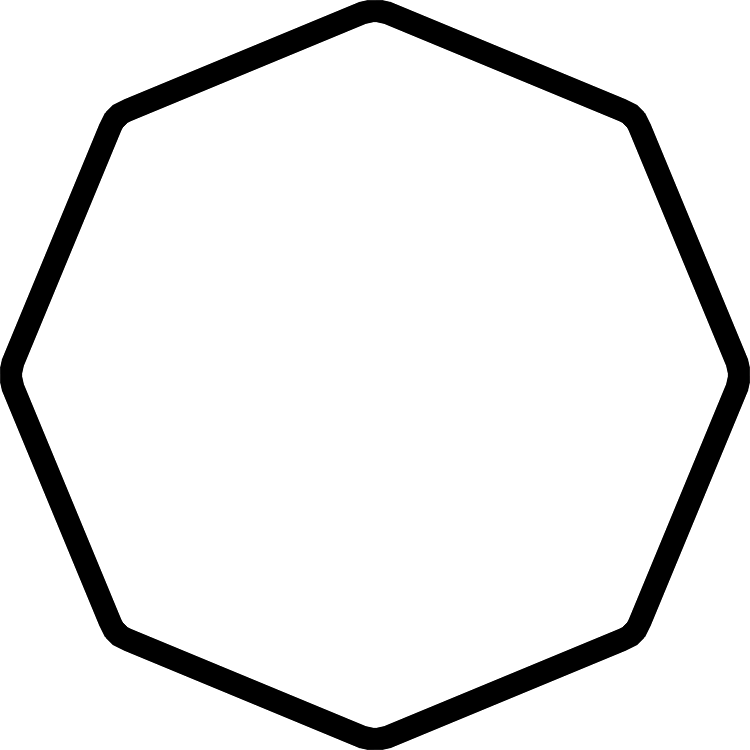}
    \caption{}
  \end{subfigure}
  \hfill
  \begin{subfigure}[b]{0.2\textwidth}
    \includegraphics[width=\textwidth]{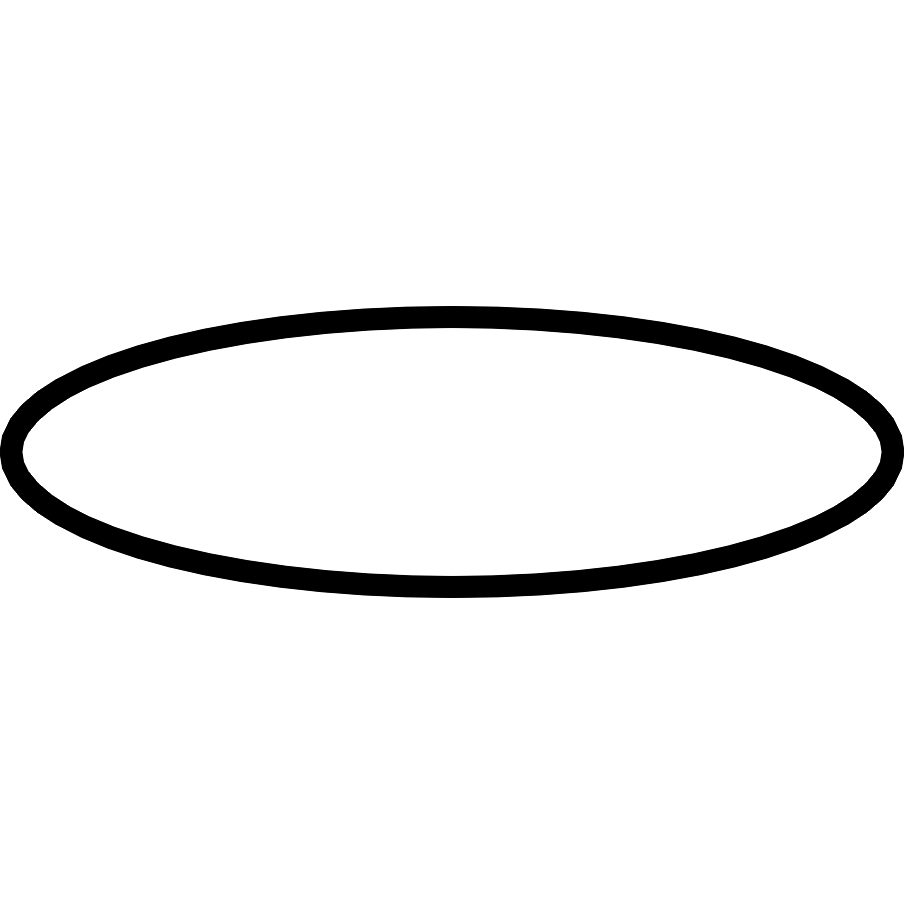}
    \caption{}
  \end{subfigure}
  \hfill
  \begin{subfigure}[b]{0.2\textwidth}
    \includegraphics[width=\textwidth]{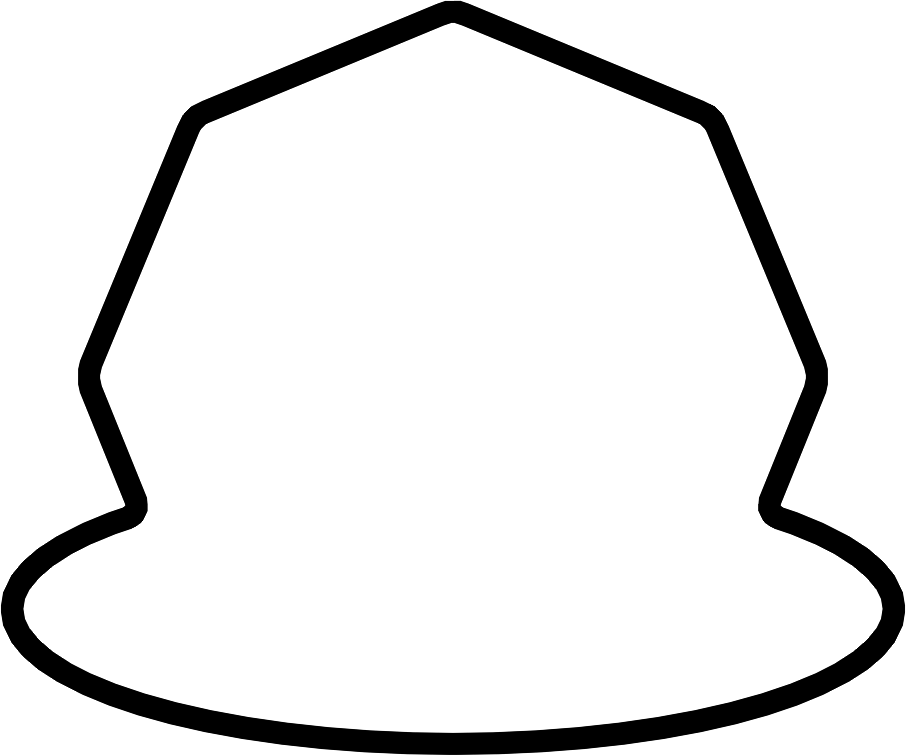}
    \caption{}
  \end{subfigure}
  \caption{Some 2D porous geometries: (a--c) from training set, (d) unseen composite.}
  \label{fig:2d-geometries}
\end{figure}

To evaluate performance in mixed porous/fluid domains we designed several 2D benchmarks inspired by~\cite{kashefi2022physics}, replacing solid inclusions with porous ones (Fig. \ref{fig:2d-geometries}).

\subsubsection{Manufactured solution test.}
As verification step we used the \emph{method of manufactured solutions} (MMS), prescribing artificial velocity/pressure fields and deriving forcing and boundary terms so that the pair is an exact solution. The dataset thus generated allows direct error evaluation of the PIPN framework. With added Darcy–Forchheimer terms, the forcing terms read
\begin{equation}\label{eq:manufactured}
  f_x = 2\mu u_x + \chi_p u_x\!\left( \mu D + \tfrac{1}{2}\rho F \lvert u\rvert\right), \qquad
  f_y = -2\mu u_y + \chi_p u_y\!\left( \mu D + \tfrac{1}{2}\rho F \lvert u\rvert\right),
\end{equation}
with $u_x=\cos(x)\sin(y)$, $u_y=-\cos(y)\sin(x)$. This provides a more demanding reference than the standard Navier--Stokes case.

\subsubsection{Case studies.}
\begin{figure}[tb]
  \centering
  \begin{subfigure}[b]{0.45\textwidth}
    \includegraphics[width=0.99\textwidth]{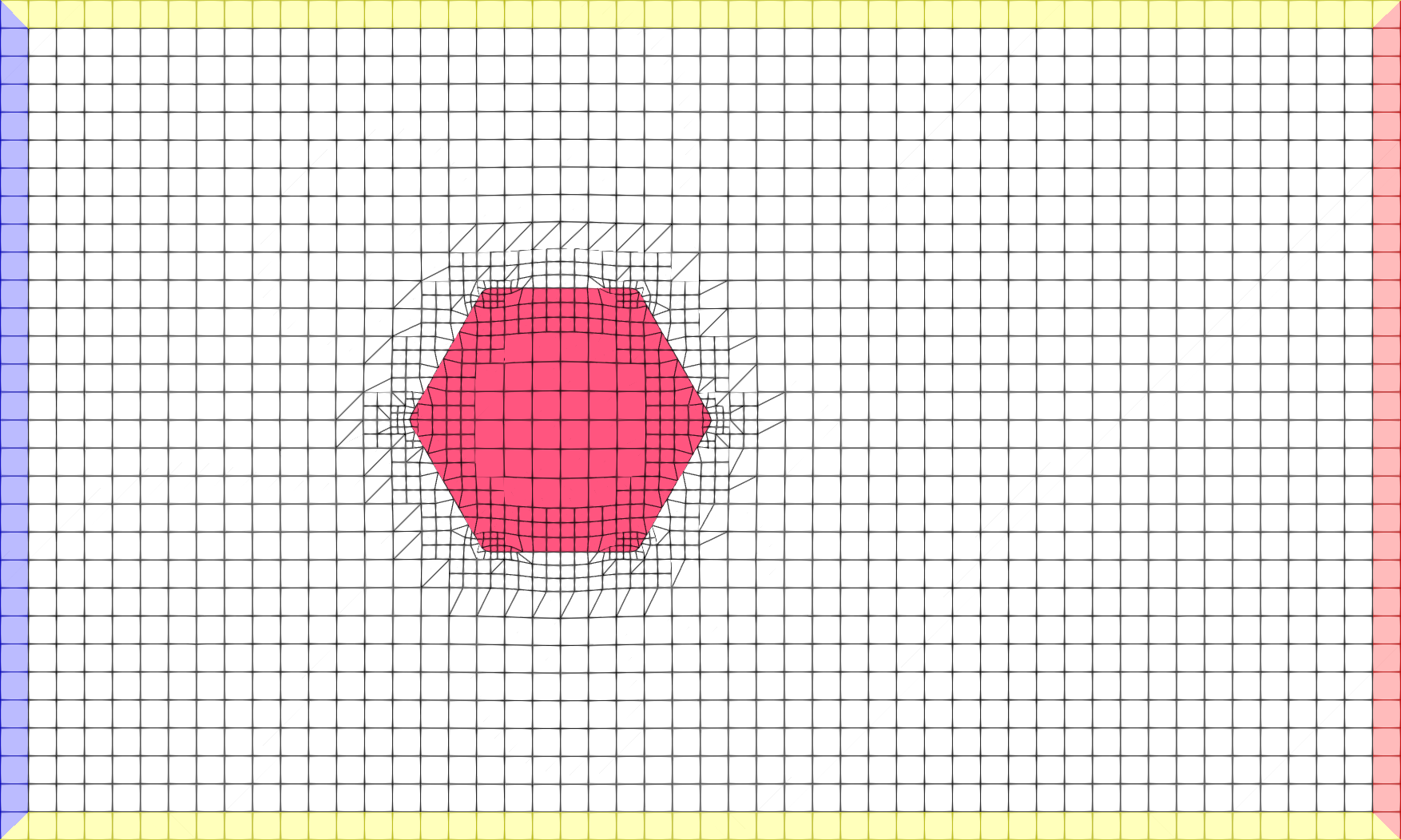}
    \caption{}
  \end{subfigure}
  \hfill
  \begin{subfigure}[b]{0.45\textwidth}
    \includegraphics[width=0.99\textwidth]{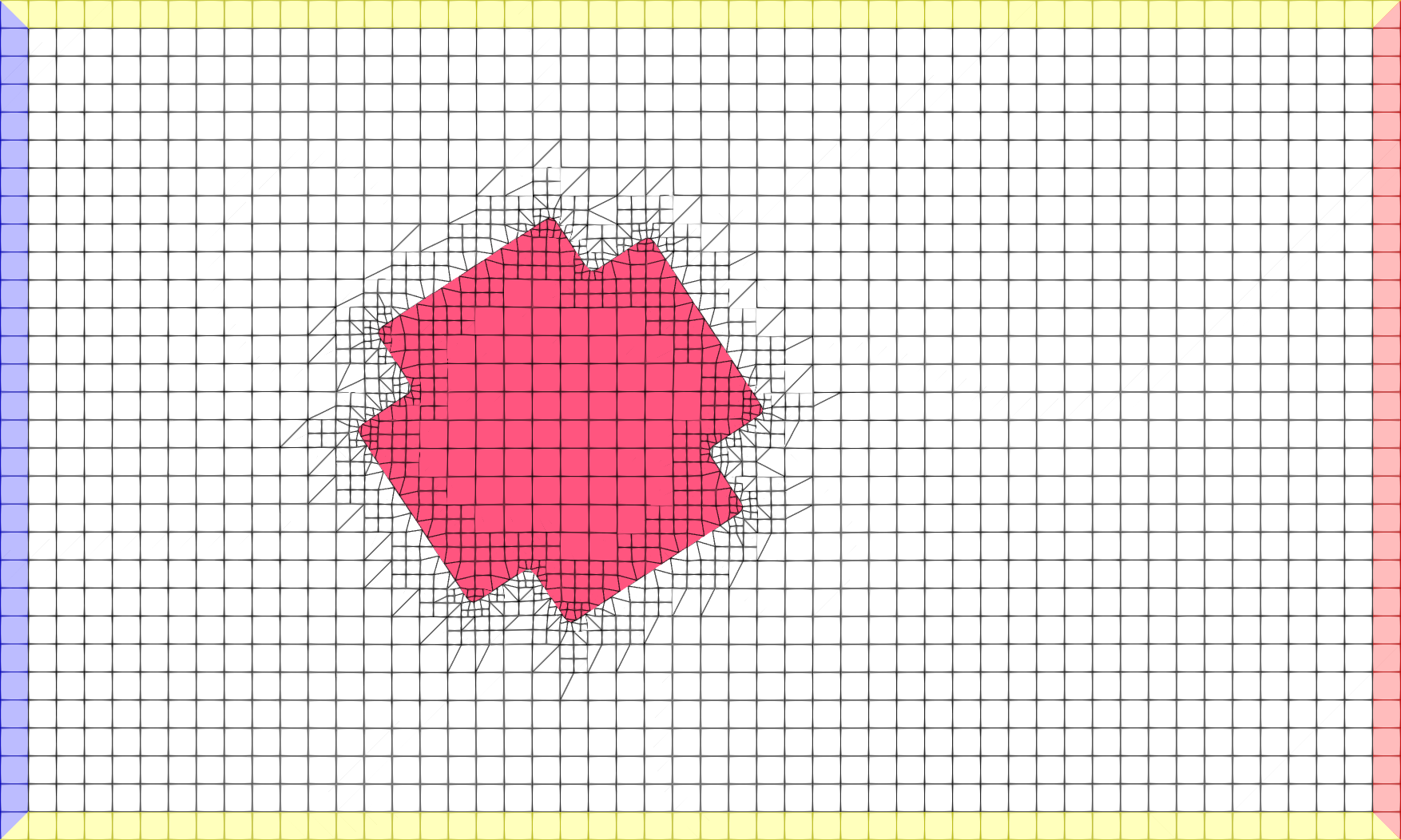}
    \caption{}
  \end{subfigure}
  \caption{2D domains generated with \texttt{snappyHexMesh}, boundary conditions: \textcolor{blue}{inlet}, \textcolor{red}{outlet}, \textcolor{orange}{wall}.}
  \label{fig:geometric}
\end{figure}

Three classes of 2D experiments were considered:
\begin{itemize}
  \item \textbf{Case 1:} baseline training using physics-informed and boundary losses only, with fixed $D$ and $F$.
  \item \textbf{Case 2:} same geometries with different boundary conditions, simulating a porous object in an oil-filled duct. Parameters (viscosity, velocity, $D$, $F$) were chosen to yield stationary wake eddies. Data augmentation included shape scaling and rotations (Fig.~\ref{fig:geometric}).
  \item \textbf{Case 3:} variable inlet velocities, angles and porosity coefficients ($D,F$), with additional augmentations (translations, offsets, random perturbations) to test generalization to unseen shapes and conditions.
\end{itemize}

\subsection{3D study cases}
\begin{figure}[tb]
  \centering
  \begin{subfigure}[b]{0.45\textwidth}
    \includegraphics[width=\textwidth]{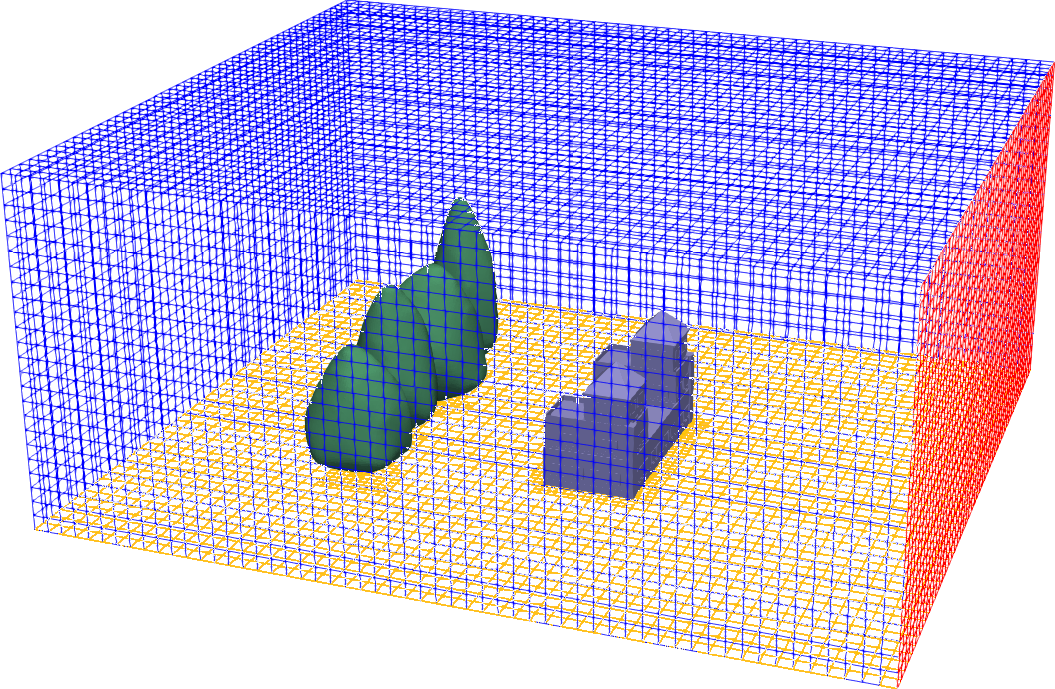}
    \caption{Pine}
    \label{fig:pine}
  \end{subfigure}
  \hfill
  \begin{subfigure}[b]{0.45\textwidth}
    \includegraphics[width=\textwidth]{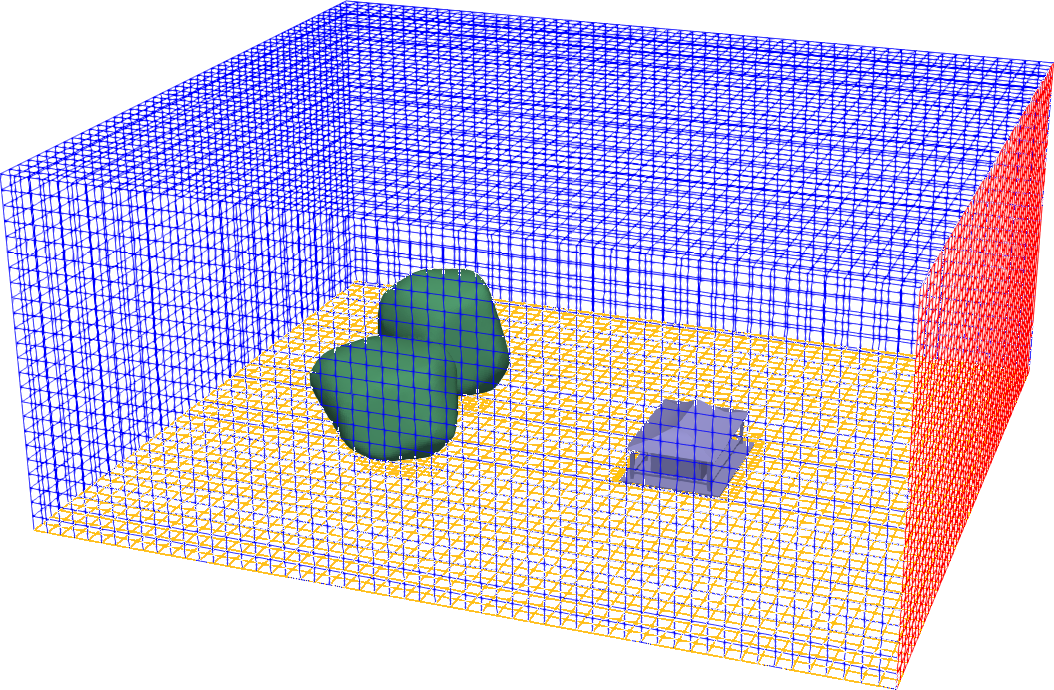}
    \caption{Acacia}
    \label{fig:acacia}
  \end{subfigure}
  \caption{3D domains with tree canopies, boundary conditions: \textcolor{blue}{inlet}, \textcolor{red}{outlet}, \textcolor{orange}{wall}.}
\end{figure}

We extended to 3D domains inspired by windbreak studies~\cite{mahgoub2021numerical}. Porous geometries representing tree canopies were generated from 3D models in Blender; convex hulls of single trees were arranged into rows, combined with house models \cite{peralta2020next} to increase complexity. Porosity was estimated from canopy images ($\phi=N_p/N$) and Darcy/Forchheimer coefficients computed via Eq.~\ref{eq:DandF}, using leaf length as $d_p$~\cite{wurTreeDatabase}. The tree species included were oak, pine, cypress, willow, acacia and eucalyptus.
The kinematic viscosity was set to $14.61\cdot10^{-6}$ m$^2$/s (air at $15^\circ$C), and inlet velocities ranged $[4\cdot10^{-6},\,10^{-5}]$ m/s, ensuring laminar regime. Randomization of tree sizes and positions produced sufficient variability (Fig.~\ref{fig:pine}--\ref{fig:acacia}).

For all experiments, data were split into training/validation/test with 60/20/20 ratio.


\section{Numerical Data Generation}
\subsection{CFD Simulations in OpenFOAM}
Training data were generated with the open-source CFD toolbox \emph{OpenFOAM}.
Given the assumptions of steady, incompressible, laminar, Newtonian flow, we employed the \texttt{simpleFoam} solver~\cite{10.1007/BFb0112677}, a finite-volume method based on the SIMPLE algorithm. Convergence thresholds were set to $10^{-4}$ and $10^{-3}$ for velocity and pressure, respectively.
Mesh discretization was performed using the \texttt{snappyHexMesh} utility, starting from a base hexahedral grid and refining around porous obstacles and interfaces to improve accuracy.
Examples of meshed domains are shown in Fig.~\ref{fig:geometric}.
Boundary conditions in OpenFOAM were set as follows: velocity (fixedValue at inlet, inletOutlet at outlet, slip at walls) and pressure (zeroGradient at inlet/walls, fixedValue at outlet).

\subsection{Point Cloud Representation}
Since PointNet-based models require point clouds, OpenFOAM cell and face centers were converted into points annotated with velocity and pressure values. A binary flag identifies porous vs.\ free-fluid regions, with associated porosity coefficients.
To ensure a fixed input size, the point clouds were uniformly sampled (preserving higher density near interfaces). Table~\ref{tab:training-params} reports the sampling sizes for each experiment.

\begin{table}[tb]
  \centering
  \caption{Training parameters of each experiment.}
  \begin{tabular}{llllllll}
    \toprule
    Experiment &  $\lambda_b$ & $\lambda_m$,$\lambda_c$ & $\lambda_d$ & $\alpha$ &  Internal & Boundary & Observations\\
    \midrule
    MMS & 1 & 1 & --- & 0.9995 & 667 & 168 & ---\\
    Duct fixed, variable & 1 & 1& 100 & 0.9990 & 1000 & 200 & 500\\
    Windbreak & 1 & 10 & 1 & 0.9990 & 2000 & 1000 & 1000\\
    \bottomrule
  \end{tabular}
  \label{tab:training-params}
\end{table}

Features were normalized: binary indices and porosity coefficients to $[0,1]$, all other variables standardized with Z-score. Since the PDE residuals are computed on normalized values, the Navier--Stokes–Darcy equations were rescaled accordingly (cf.~\cite{xu2025preprocessing}). For example, Eq. \ref{eq:cont-steady} in 2D becomes
\begin{equation*}
  \frac{\sigma_{u_x}}{\sigma_x}\frac{\partial u_x}{\partial x}+
  \frac{\sigma_{u_y}}{\sigma_y}\frac{\partial u_y}{\partial y}=0
\end{equation*}
with analogous scaling applied to the momentum equation. This ensures consistency between the standardized training data and the physics-informed loss.


\section{Physics-Informed  and Geometry-Aware ANNs}\label{sec:PIPN}

\subsection{Physics-Informed Neural Networks}

Physics-Informed Neural Networks (PINNs) solve PDEs by minimizing a loss function that combines agreement with available data and consistency with the governing equations.
For the steady incompressible Navier--Stokes--Darcy equations (\ref{eq:forchheimer})
the PINN loss function takes the form:

\begin{equation*}
  \mathcal{L}(\theta) = \lambda_m\mathcal{L}_m(\theta) + \lambda_c\mathcal{L}_c(\theta) + \lambda_b\mathcal{L}_b(\theta) + \lambda_d\mathcal{L}_d(\theta)
\end{equation*}

\begin{equation*}
  \mathcal{L}_c(\theta)  =\frac{1}{N_f} \sum_{i=1}^{N_f}  \left\| \nabla \cdot \mathbf{u_i^\theta}  \right\|^2
\end{equation*}

\begin{equation*}
  \mathcal{L}_m(\theta)  =\frac{1}{N_f} \sum_{i=1}^{N_f}  \left\| \rho\, (\mathbf{u_i^\theta} \cdot \nabla)\mathbf{u_i^\theta} + \nabla p_i^\theta - \mu \nabla^2 \mathbf{u_i^\theta} +  \chi_{\Omega_p}\Big ( \mu D + \frac{1}{2}\rho F \lvert \mathbf{u_i^\theta}\rvert \Big )\mathbf{u_i^\theta}  \right\|^2
\end{equation*}

\begin{equation*}
  \mathcal{L}_b(\theta)  = \frac{1}{N_b} \sum_{i=1}^{N_b}  \left\|\mathbf{u_i^\theta}  - \mathbf{u_i}\right\|^2 +  \left\|p_i^\theta  - p_i\right\|^2 \quad
  \mathcal{L}_d(\theta)  =  \frac{1}{N_d} \sum_{i=1}^{N_d}  \left\|\mathbf{u_i^\theta}  - \mathbf{u_i}\right\|^2 +  \left\|p_i^\theta  - p_i\right\|^2
\end{equation*}

Here:
$N_d$ is the number of data points where measurements or simulation data are available;
$N_f$ is the number of collocation points used to enforce the PDE;
$N_b,N_f$ is the number of collocation points used to enforce the boundary conditions and equations respectively;
$\theta$ are the network parameters.
$\chi_{\Omega_p}$ is $1$ if the collocation point is inside the porous domain, $0$ otherwise. This formulation allows to remove the porous term in the fluid region.
$\lambda_d,\lambda_f,\lambda_b$ are parameters which controls the importance of each loss term.
$\mathcal{L}_d$ ensures agreement with available data;
$\mathcal{L}_c,\mathcal{L}_m$ penalize violations of the governing equations using automatic differentiation to compute derivatives;
$\mathcal{L}_b$ term enforces that the boundary conditions.
Automatic differentiation is a technique for evaluating exact derivatives of a computational graph by systematically applying the chain rule,
allowing gradients of PDE residuals to be obtained with machine precision and without symbolic derivation.

\subsection{Physics-informed PointNets}

A key point is that a PINN is trained on a single geometry and cannot directly generalize to new shapes: this requires retraining for each case. Such limitation makes conventional PINNs impractical for exploring a wide range of geometric configurations in industrial design.

Physics-Informed PointNets (PIPNs) extend this concept to unstructured geometric domains by combining the PINN framework with the PointNet architecture, which processes 3D or 2D point clouds in a permutation-invariant way.

In PIPN, the input is a point cloud representing both the geometry and the associated physical coordinates, and the network outputs the predicted fields. The physics-informed loss is then applied to these outputs at the point level, enabling the model to generalize across different domain shapes without remeshing or reparameterization.

Compared to the original PIPN formulation, our implementation incorporates porous-media-specific terms in the PDE residuals, allowing the same framework to handle both Navier–Stokes and Darcy-modified equations within their respective subdomains.
Furthermore, the approach is extended to fully 3D problems, requiring modifications in the PointNet architecture and the derivative computation pipeline to handle 3D spatial gradients efficiently.

\subsubsection{Architecture}
The first step of this work was to extend the existing PIPN architecture \cite{kashefi2022physics} to take into account the porous medium. This proposed model is based on PointNets\cite{qi2017pointnet} but adds the physics loss to the training process. This involved two main challenges: the porous zone had to be input to the model and the losses had to be modified to disable the source term if the collocation point was not porous. Therefore the final implementation added a  Signed Distance Function (SDF) to the input points and a binary indicator feature $ID$ to the momentum losses, corresponding to $\chi_{\Omega_p}$ in $\mathcal{L_m}$.
This allows the source term to be disabled if a collocation point is inside the fluid region and at the same time ensures the correctness of the automated differentiation process.
Figure \ref{fig:pipn-porous} shows a diagram of such architecture with the associated input dimension of each layer. As opposed to the original PIPN implementation the input included the position of each point $X \in R^d$ the SDF function associated to the point and a one hot encoded feature representing the type of boundary condition ($Id_b$). To improve the the global embedding the SDF and boundary id were concatenated to the output of the first shared layers which is then used to generate the global embedding representation of the domain. This allows the global embedding to incorporate both positional information and porosity information. The rest of the architecture is the same as the original PIPN implementation.

\subsection{Physics-Informed Geometry-Aware Neural Operator}
While PIPN processes each geometry as an independent point cloud and learns a mapping from coordinates and physical parameters to flow fields, it still operates on fixed boundary conditions on all samples.

PI-GANO \cite{zhong2025pigano} extends this concept by embedding both the geometry and variable boundary conditions into a neural operator framework, which learns a mapping between parametrized function spaces. This is achieved by representing the geometry in a latent space and the boundary conditions into another latent space. The two representations, together with the permutation invariant representation of each point are then combined to predict the behavior of the physical system.

\subsubsection{Architecture} The model was implemented as per the original paper but the porous SDF and $Id_b$ features were added to the Geometry Encoding module. Moreover the sum operation before the output was replaced with a shared MLP layer. This last modification has the advantage of not using multiple branches for each variable, therefore reducing the number of operations and parameters while obtaining better performance. The variable inlet velocity $U$ and the Darcy and Forchheimer coefficients ($D$ and $F$) were added as inputs to the branch module together with the positions at which the boundary condition was available. If one of the conditions was not available at that position, its value was set to 0. Finally the $ID$ feature was added to the momentum losses as in the PIPN. To take advantage of the mini-batching mechanism the number of points $M$ input to the branch was fixed, using the ratio of the minimum number of points of each boundary to the number of total boundary samples.

\begin{figure}[tb]
  \centering
  \begin{subfigure}{0.36\linewidth}
    \centering
    \includegraphics[width=0.39\linewidth]{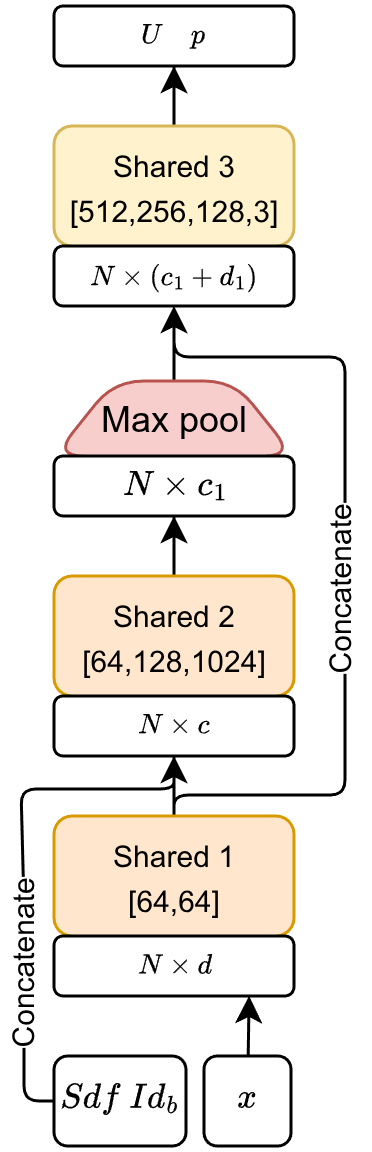}
    \caption{The PIPN architecture}
    \label{fig:pipn-porous}
  \end{subfigure}
  \hfill
  \begin{subfigure}{0.6\linewidth}
    \centering
    \includegraphics[angle=-90,width=\linewidth]{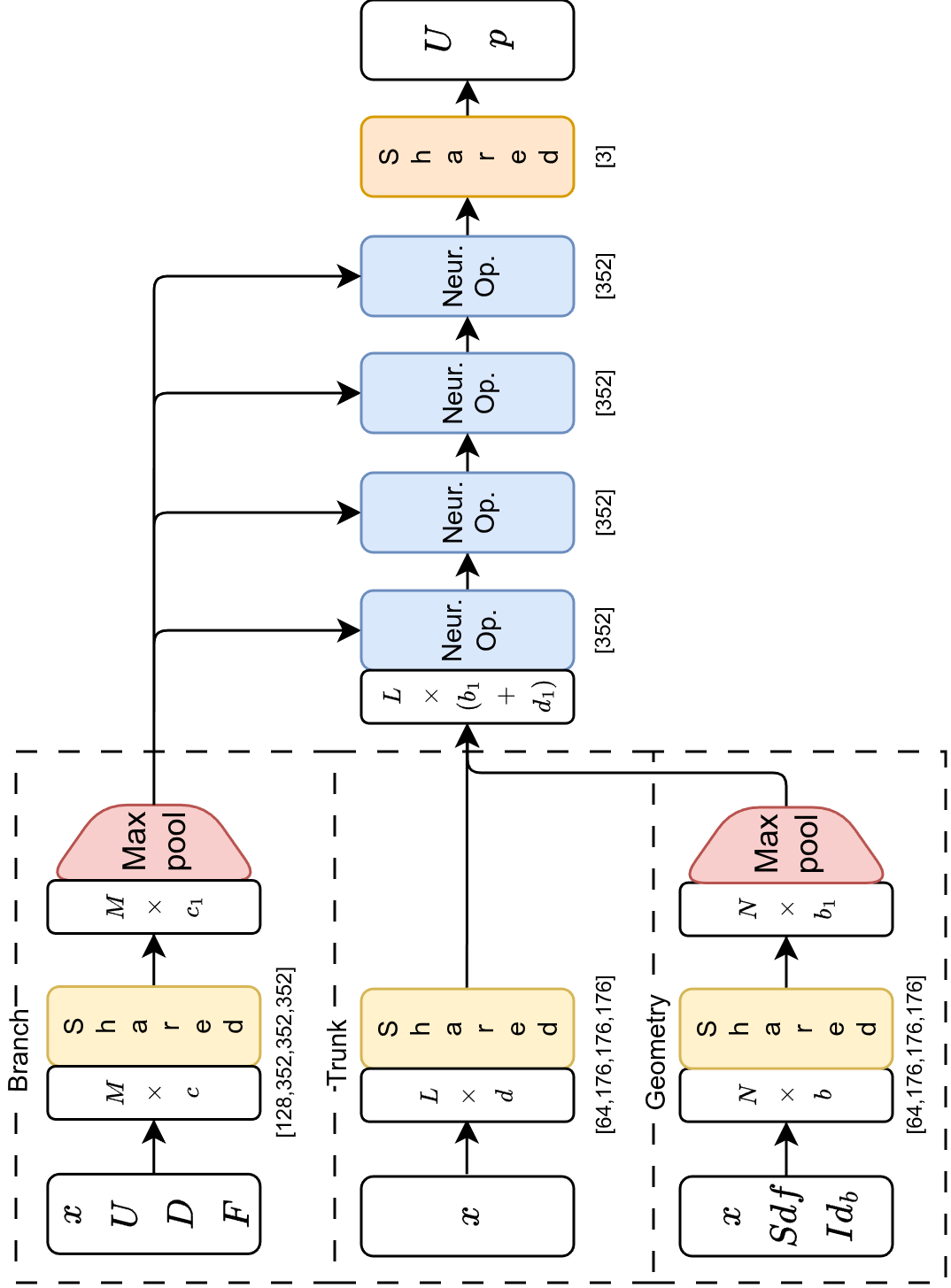}
    \caption{The PI-GANO architecture}
    \label{fig:pi-gano-porous}
  \end{subfigure}
  \caption{Architectures modified to take into account porous medium.}
  \label{fig:arch-porous}
\end{figure}

\subsection{Training}
All the models were trained using the SiLU \cite{elfwing2018sigmoid} activation function while in the manufactured solutions experiment the hyperbolic tangent was used. In all experiments the Adam optimizer with a learning rate of $0.001$ and an exponential learning rate decay $\alpha$ was used. In all experiments but the manufactured solutions one a dropout layer with $p=0.05$ was added to the last two layers before the output to prevent overfitting. All model were trained for $3000$ epochs to allow  convergence and the loss coefficients were set according to Table \ref{tab:training-params}.
Those values allowed to obtain accurate results and were set through experimentation on different cases. It must be noted that the final values are consistent with \cite{harmening2024data}.


\section{Experimental Results}\label{sec:results}

\subsection{Case 1: Manufactured solutions}
We use the PIPN of Fig.~\ref{fig:pipn-porous}.
Performance, reported as MAE, is comparable to~\cite{kashefi2022physics} on both training and unseen geometries (Table~\ref{tab:manufactured-pipn-error}).

\begin{table}[tb]
  \centering
  \begin{tabular}{lllllll}
    \toprule
    \multicolumn{1}{c}{ } & \multicolumn{3}{c}{Training}& \multicolumn{3}{c}{Unseen}\\
    \cmidrule(r){2-4} \cmidrule(r){5-7}
    & Global & Porous region & Fluid region& Global & Porous region & Fluid region\\
    \midrule
    $u_x$ & $1.20 \cdot 10^{-2}$ &  $1.20\cdot 10^{-2}$ &$1.20 \cdot 10^{-2}$ &
    $1.23 \cdot 10^{-2}$ &  $1.23 \cdot 10^{-2}$ &$1.23 \cdot 10^{-2}$ \\

    $u_y$ & $9.39 \cdot 10^{-3}$ &$8.13 \cdot 10^{-3}$ & $\mathbf{9.63 \cdot 10^{-3}}$&
    $9.38 \cdot 10^{-3}$ &  $8.14 \cdot 10^{-3}$ &$\mathbf{9.61 \cdot 10^{-3}}$ \\

    $p$ & $1 \cdot 10^{-2}$ &  $8.93 \cdot 10^{-3}$ & $\mathbf{1.02 \cdot 10^{-2}}$&
    $1.02 \cdot 10^{-2}$ &  $8.73 \cdot 10^{-3}$ &$\mathbf{1.05 \cdot 10^{-2}}$ \\
    \bottomrule
  \end{tabular}
  \caption{MMS errors of the PIPN predictions.}
  \label{tab:manufactured-pipn-error}
\end{table}

\begin{figure}[tb]
  \centering
  \begin{subfigure}[b]{0.32\textwidth}
    \includegraphics[width=0.99\textwidth]{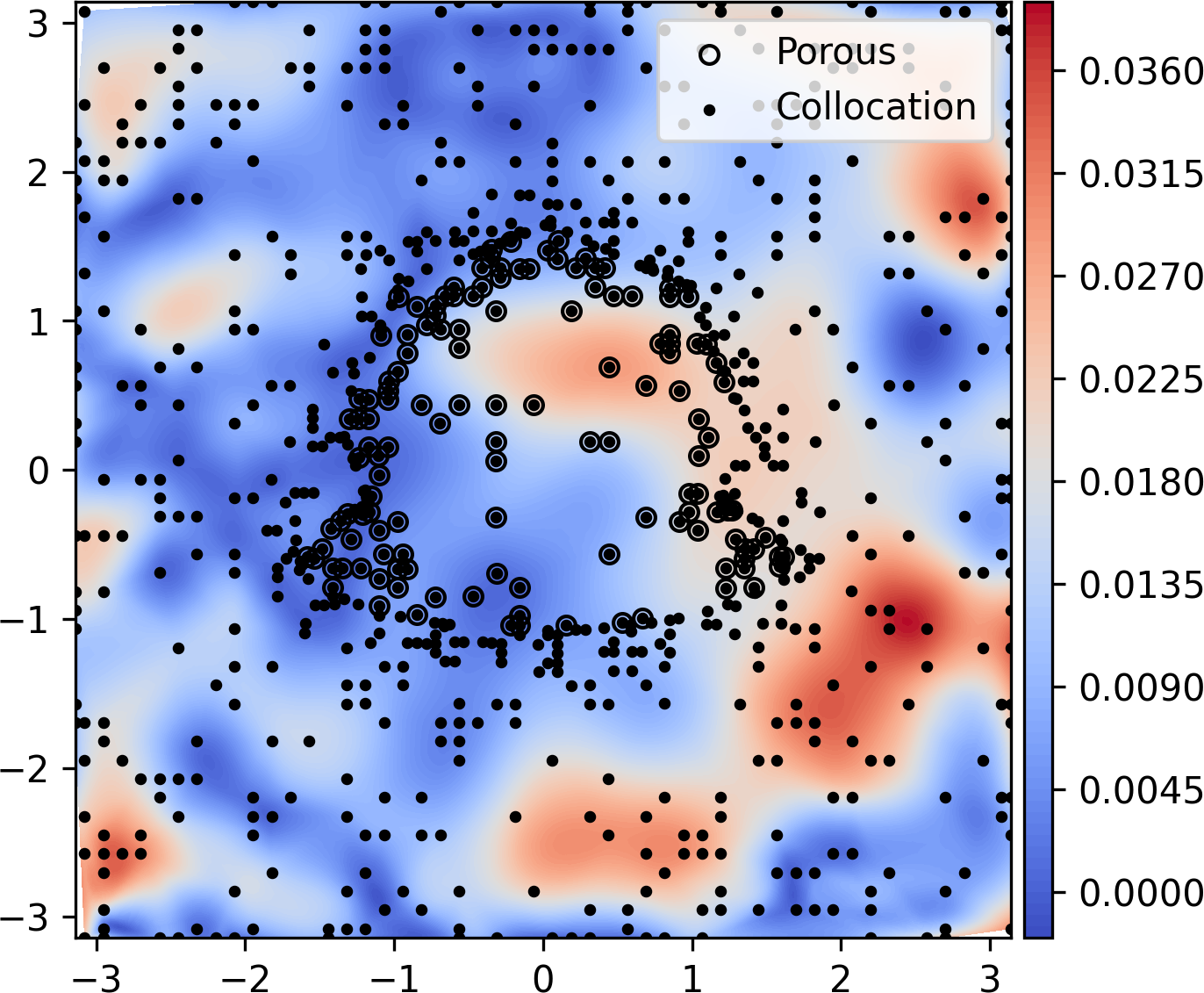}
    \caption{$u_x$}
  \end{subfigure}
  \begin{subfigure}[b]{0.32\textwidth}
    \includegraphics[width=0.99\textwidth]{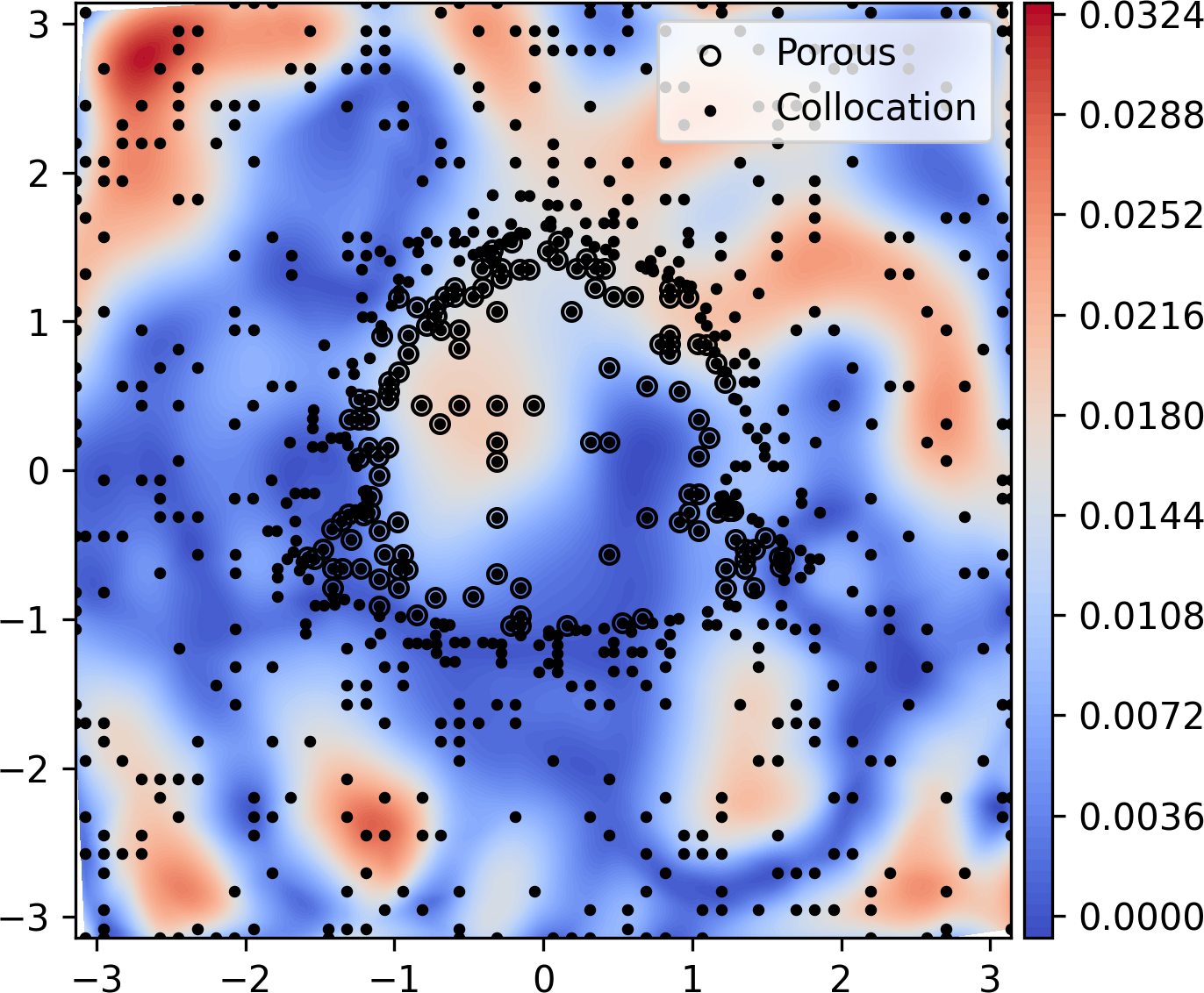}
    \caption{$u_y$}
  \end{subfigure}
  \begin{subfigure}[b]{0.32\textwidth}
    \includegraphics[width=0.99\textwidth]{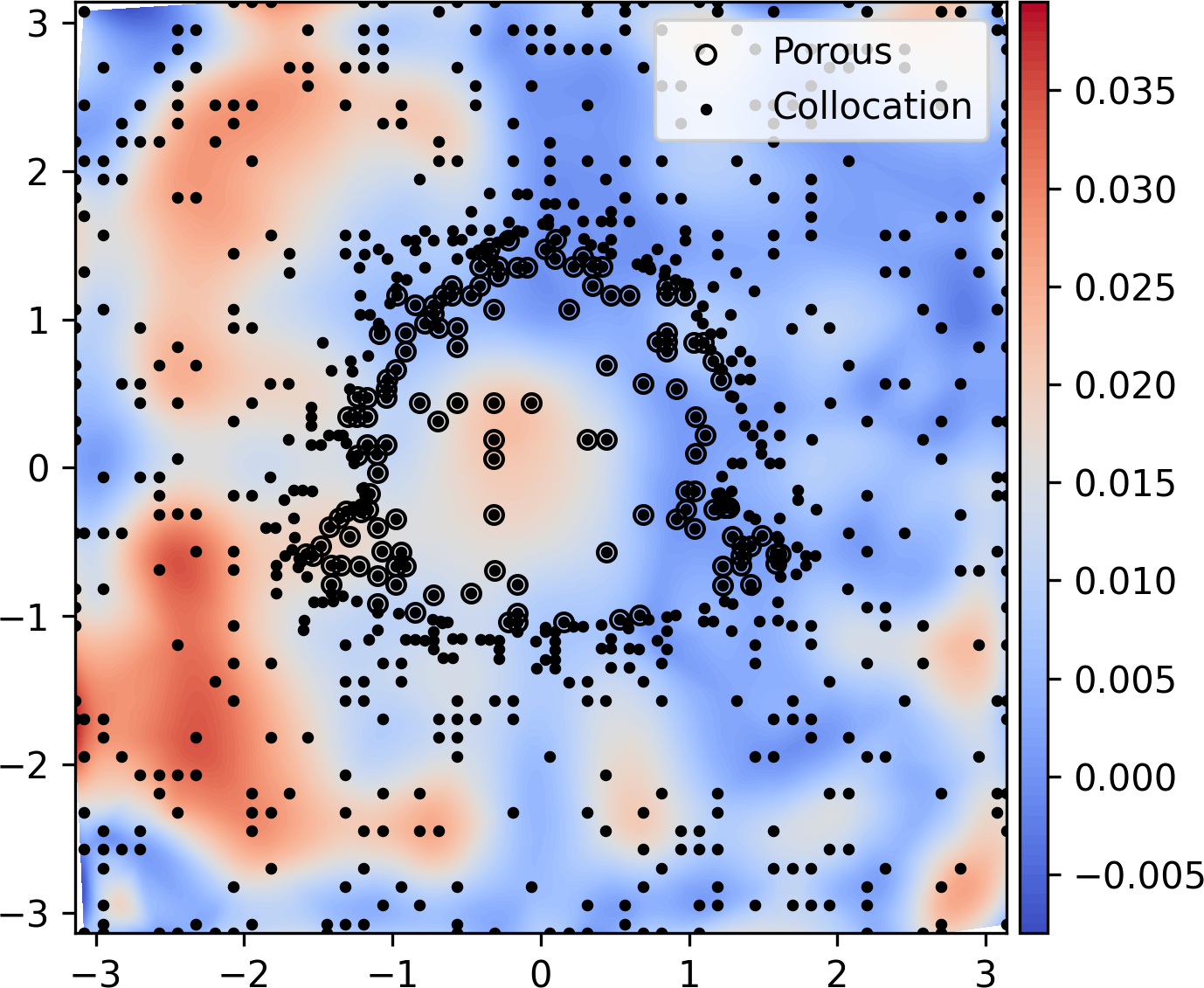}
    \caption{$p$}
  \end{subfigure}
  \caption{MAE obtained from the PIPN prediction on the unseen geometry in Fig.~\ref{fig:2d-geometries}.}
  \label{fig:pipn-manufacture-mae}
\end{figure}

Errors are uniformly low; the porous subdomain tends to be slightly easier than the free-flow region. Fig.~\ref{fig:pipn-manufacture-mae} (unseen sample) shows no structured error bias.

\subsection{Case 2: 2D porous flow with fixed boundary conditions}
We reuse the PIPN as above. Table~\ref{tab:fixed-duct-pipn-error} reports MAE on test and unseen sets.

\begin{table}[tb]
  \centering
  \begin{tabular}{lllllll}
    \toprule
    \multicolumn{1}{c}{ } & \multicolumn{3}{c}{Test}& \multicolumn{3}{c}{Unseen}\\
    \cmidrule(r){2-4} \cmidrule(r){5-7}
    & Global & Porous region & Fluid region& Global & Porous region & Fluid region\\
    \midrule
    $u_x$ & $3.1 \cdot 10^{-3}$ &  $\mathbf{3.6 \cdot 10^{-3}}$ &$3.1 \cdot 10^{-3}$ &
    $9.3 \cdot 10^{-3}$ &  $\mathbf{1.1 \cdot 10^{-2}}$ &$9.1 \cdot 10^{-3}$ \\

    $u_y$ & $1.2 \cdot 10^{-3}$ &$\mathbf{2.1 \cdot 10^{-3}}$ & $1.1 \cdot 10^{-3}$&
    $3.7 \cdot 10^{-3}$ &  $\mathbf{6.2 \cdot 10^{-3}}$ &$3.4 \cdot 10^{-3}$ \\

    $p$ & $9 \cdot 10^{-4}$ &  $\mathbf{1.7 \cdot 10^{-3}}$ & $8 \cdot 10^{-4}$&
    $6.5 \cdot 10^{-3}$ &  $\mathbf{8.8 \cdot 10^{-3}}$ &$6.2 \cdot 10^{-3}$ \\
    \bottomrule
  \end{tabular}
  \caption{MAE of the PIPN predictions with fixed boundary conditions.}
  \label{tab:fixed-duct-pipn-error}
\end{table}

\begin{figure}[tb]
  \centering
  \begin{subfigure}[b]{0.32\textwidth}
    \includegraphics[width=0.99\textwidth]{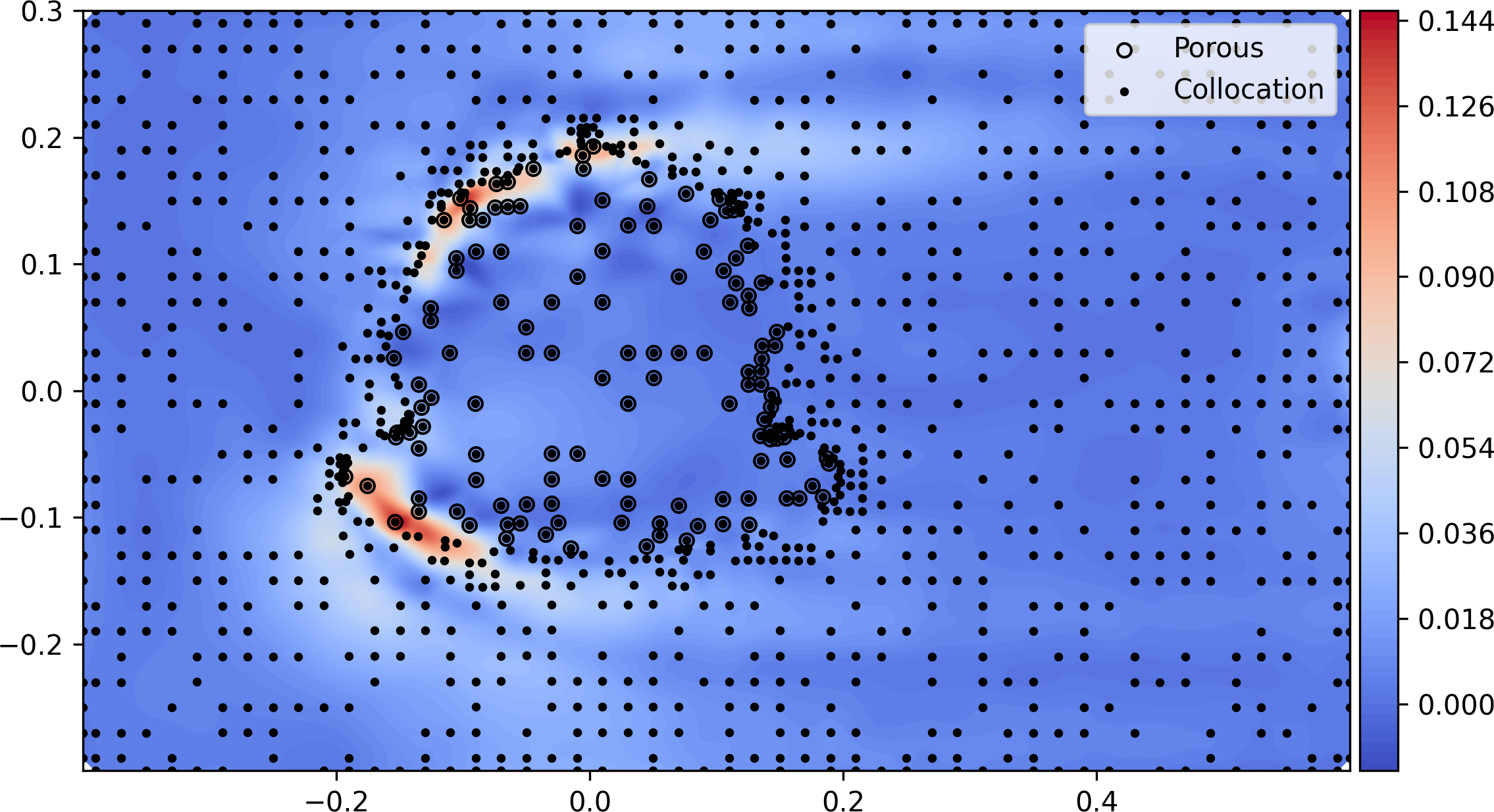}
    \caption{$U$ MAE}
  \end{subfigure}
  \begin{subfigure}[b]{0.32\textwidth}
    \includegraphics[width=0.99\textwidth]{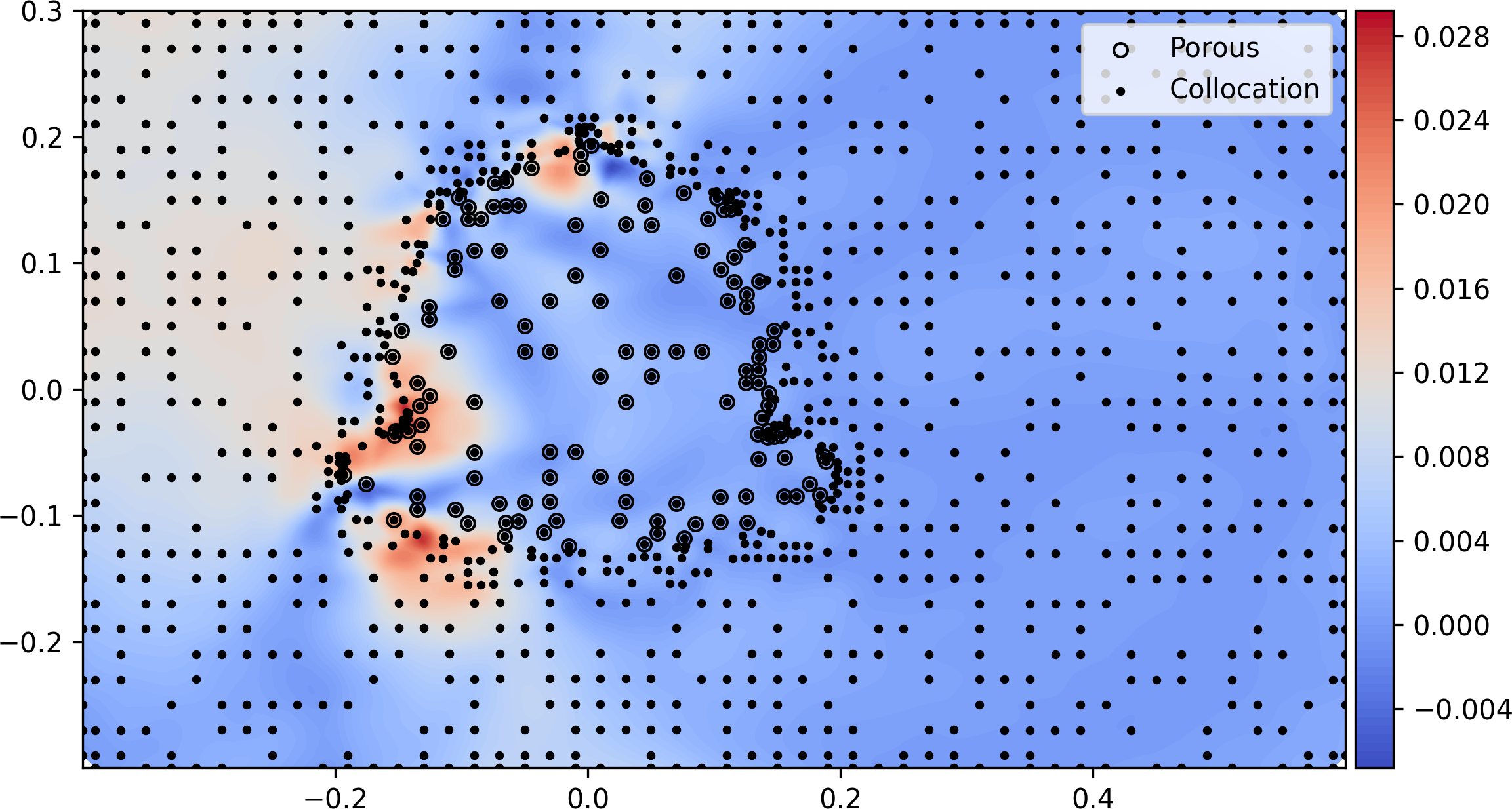}
    \caption{$p$ MAE }
  \end{subfigure}
  \begin{subfigure}[b]{0.32\textwidth}
    \includegraphics[width=0.99\textwidth]{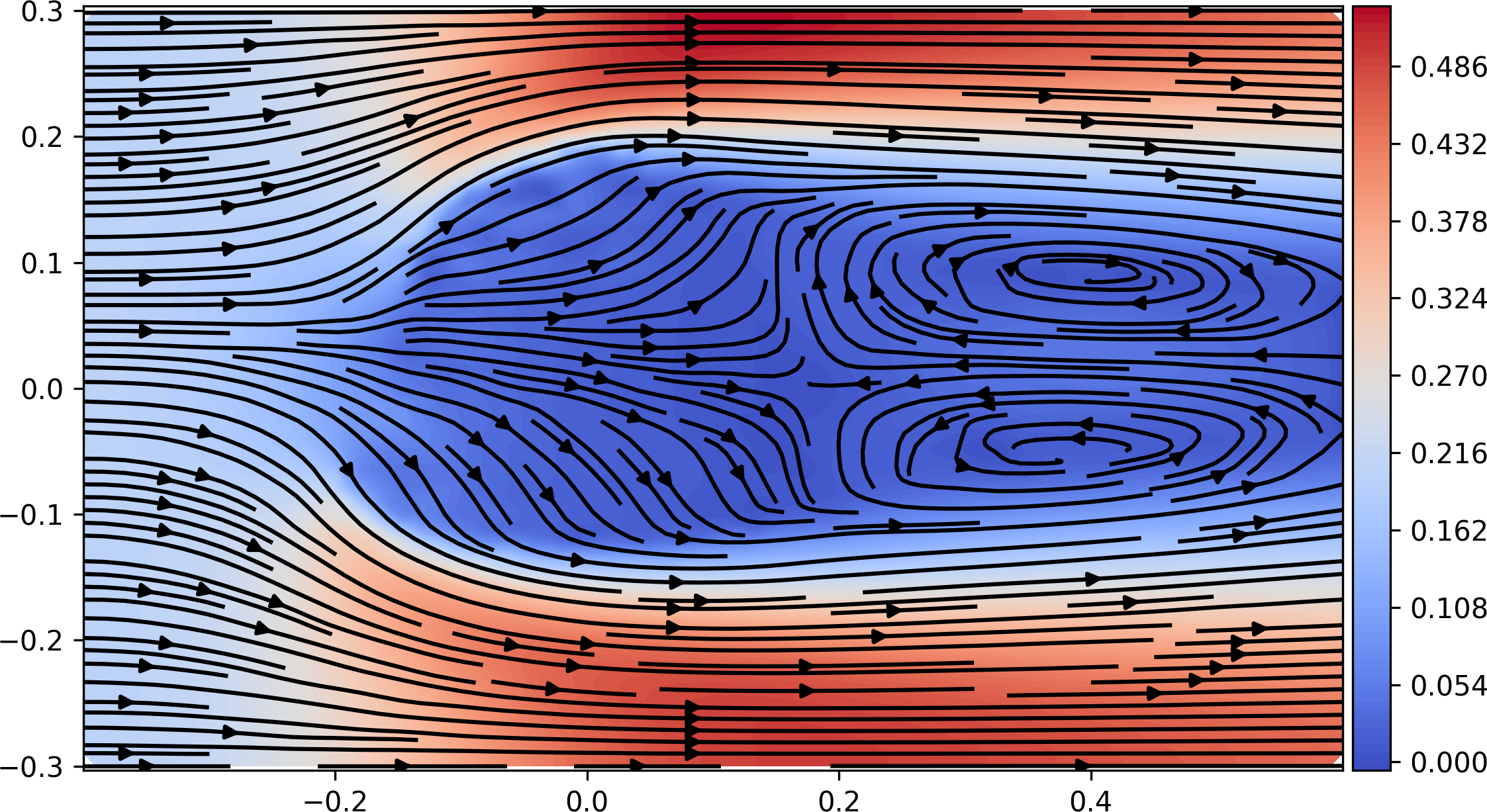}
    \caption{True $U$ }
  \end{subfigure}
  \begin{subfigure}[b]{0.32\textwidth}
    \includegraphics[width=0.99\textwidth]{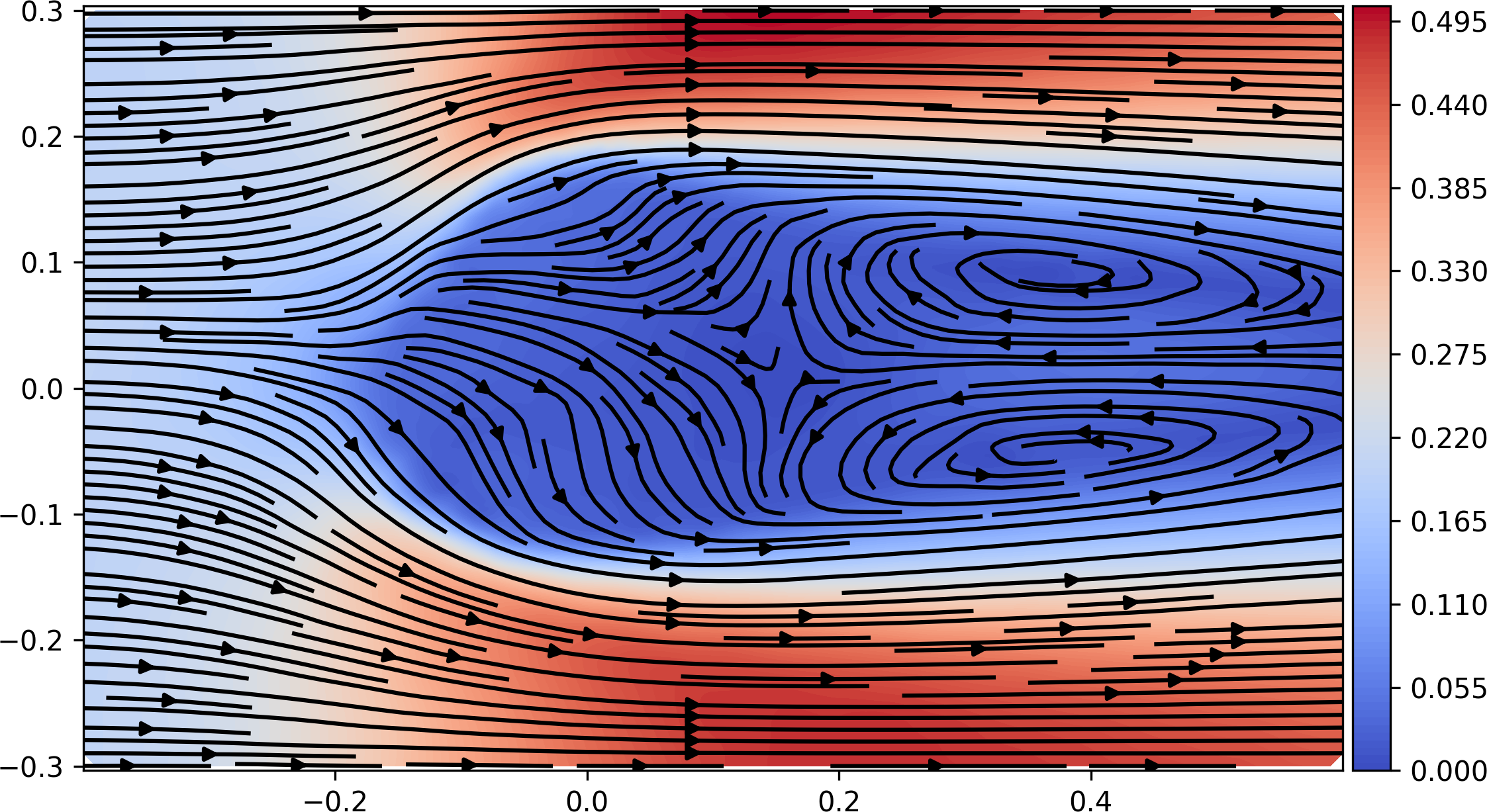}
    \caption{Predicted $U$ }
  \end{subfigure}
  \begin{subfigure}[b]{0.32\textwidth}
    \includegraphics[width=0.99\textwidth]{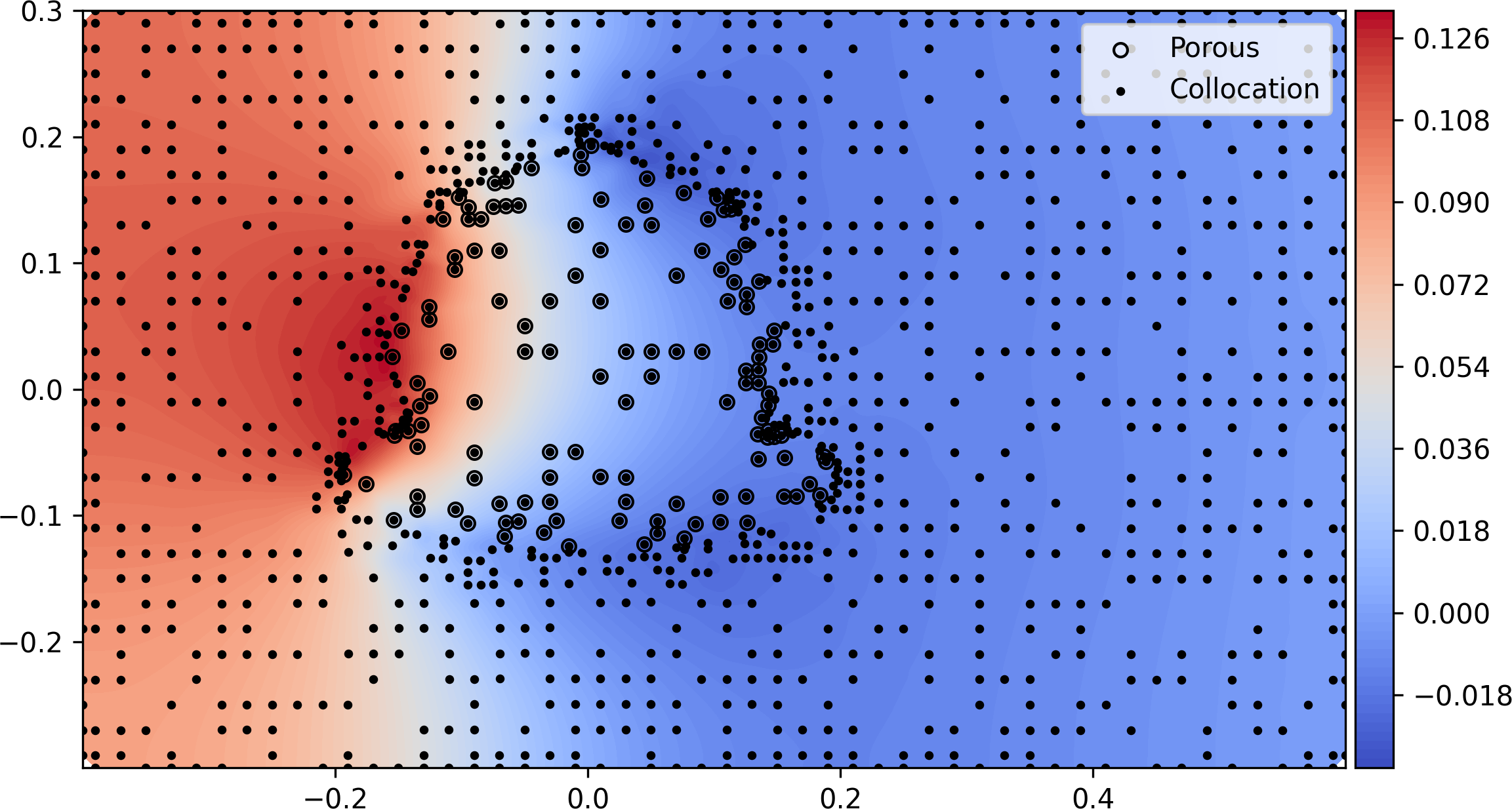}
    \caption{True $p$ }
  \end{subfigure}
  \begin{subfigure}[b]{0.32\textwidth}
    \includegraphics[width=0.99\textwidth]{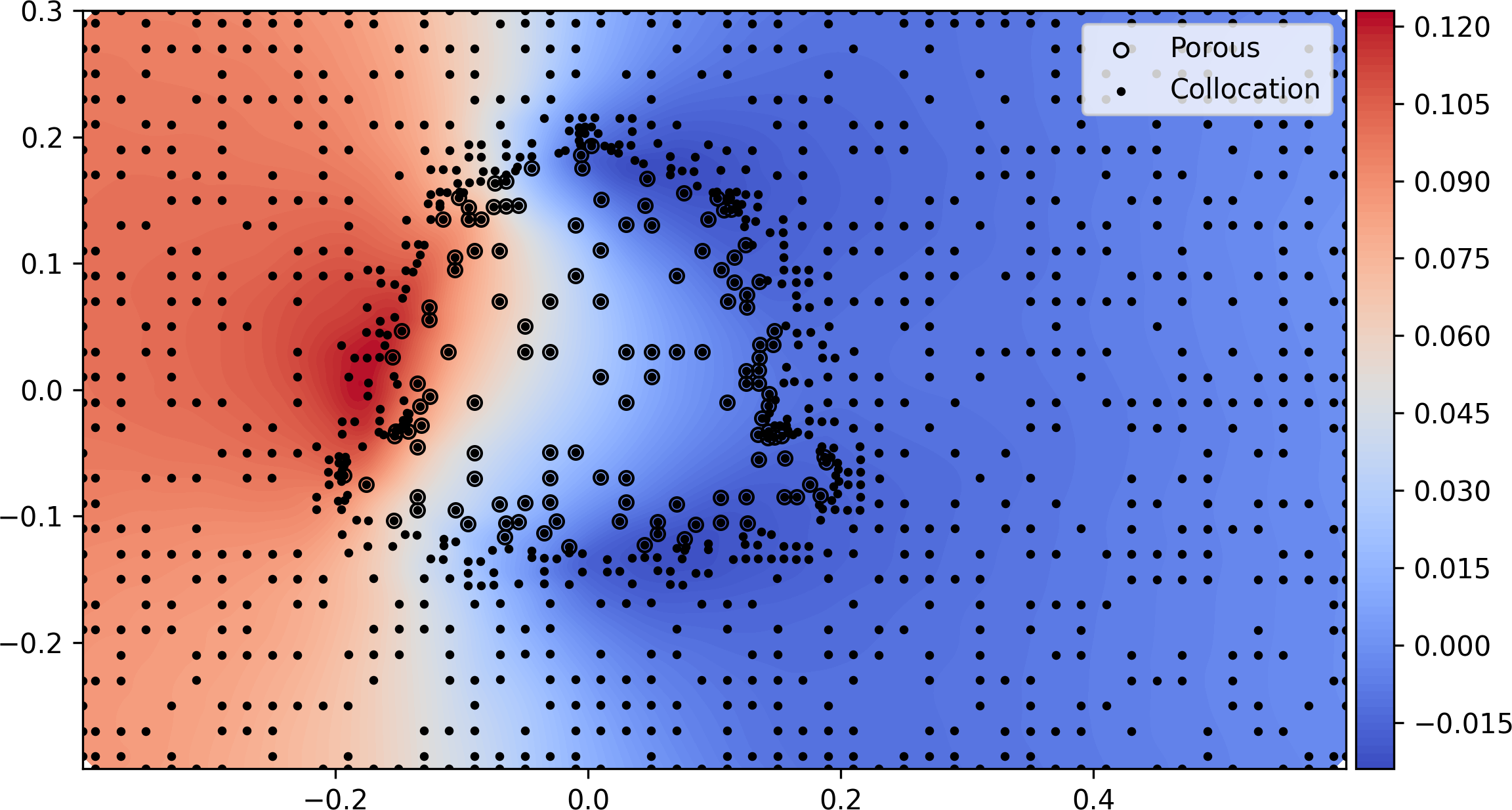}
    \caption{Predicted $p$ }
  \end{subfigure}
  \caption{Errors, prediction and ground truth on an unseen 2D geometry (cf.\ Fig.~\ref{fig:2d-geometries}).}
  \label{fig:pipn-fixed-bc-plots}
\end{figure}

Errors remain low; $u_x$ is consistently the most challenging (stronger gradients across the porous/free-flow interface). Errors are higher on the unseen set. Fig.~\ref{fig:pipn-fixed-bc-plots} (unseen geometry) illustrates error localization near sharp corners and correct wake prediction.
The average simulation time on the test set is 0.02s while OpenFOAM took 1.17s, providing a significant speedup.

\subsection{Case 3: 2D porous flow with variable boundary conditions}
We evaluate PI-GANO (larger geometry embedding; cf.\ Fig.~\ref{fig:pi-gano-porous}) on test and unseen sets with BCs/parameters beyond training ranges.

\begin{table}[tb]
  \centering
  \begin{tabular}{lllllll}
    \toprule
    \multicolumn{1}{c}{ } & \multicolumn{3}{c}{Test}& \multicolumn{3}{c}{Unseen}\\
    \cmidrule(r){2-4} \cmidrule(r){5-7}
    & Global & Porous region & Fluid region& Global & Porous region & Fluid region\\
    \midrule
    $u_x$ & $1.95 \cdot 10^{-3}$ &  $1.63 \cdot 10^{-3}$ &$\mathbf{1.99 \cdot 10^{-3}}$ &

    $1.57 \cdot 10^{-2}$ &  $\mathbf{1.90 \cdot 10^{-2}}$ &$1.53 \cdot 10^{-2}$\\

    $u_y$ & $8.47 \cdot 10^{-4}$ &  $\mathbf{9.05 \cdot 10^{-4}}$ &$8.40 \cdot 10^{-4}$ &

    $5.12 \cdot 10^{-3}$ &  $\mathbf{5.25 \cdot 10^{-3}}$ &$5.10 \cdot 10^{-3}$\\

    $p$ & $4.00 \cdot 10^{-4}$ &  $\mathbf{6.43 \cdot 10^{-4}}$ &$4.77 \cdot 10^{-4}$ &

    $6.86 \cdot 10^{-3}$ &  $\mathbf{7.35 \cdot 10^{-3}}$ &$6.80 \cdot 10^{-3}$\\
    \bottomrule
  \end{tabular}
  \caption{MAE of the PI-GANO predictions with variable boundary conditions.}
  \label{tab:variable-duct-pigano-error}
\end{table}

\begin{figure}[tb]
  \centering
  \begin{subfigure}[b]{0.32\textwidth}
    \includegraphics[width=0.99\textwidth]{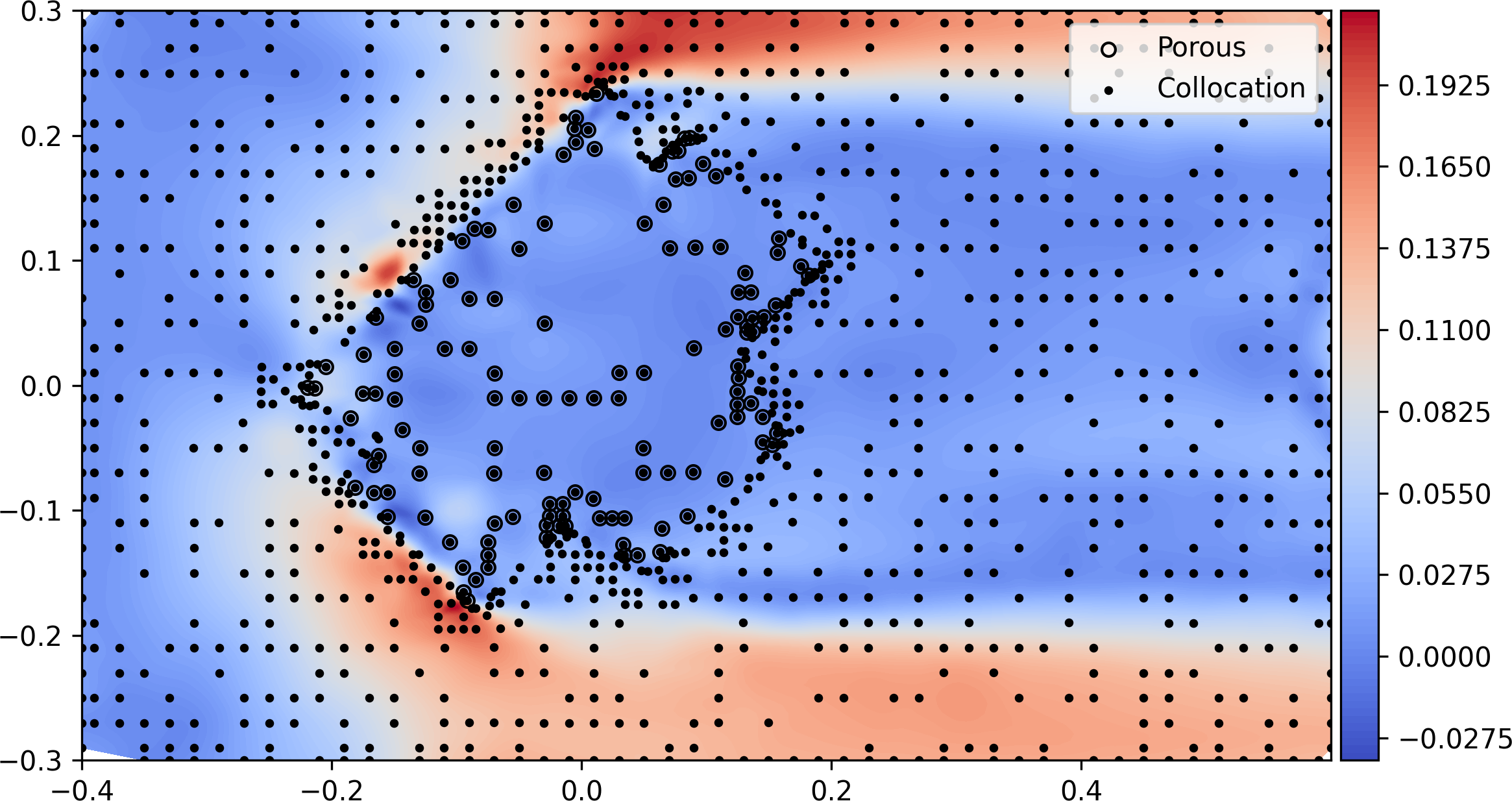}
    \caption{$U$ MAE}
  \end{subfigure}
  \begin{subfigure}[b]{0.32\textwidth}
    \includegraphics[width=0.99\textwidth]{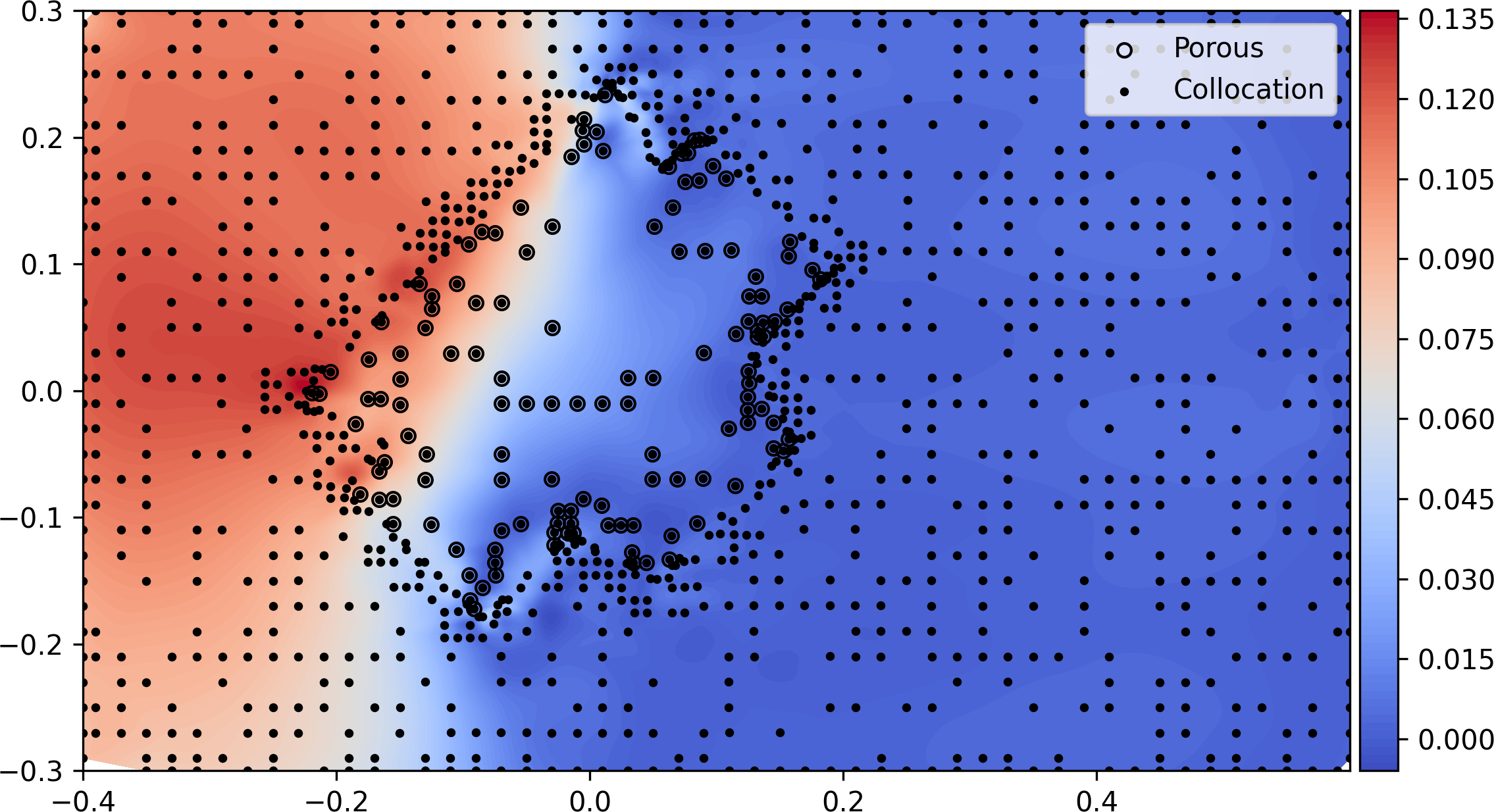}
    \caption{$p$ MAE }
  \end{subfigure}
  \begin{subfigure}[b]{0.32\textwidth}
    \includegraphics[width=0.99\textwidth]{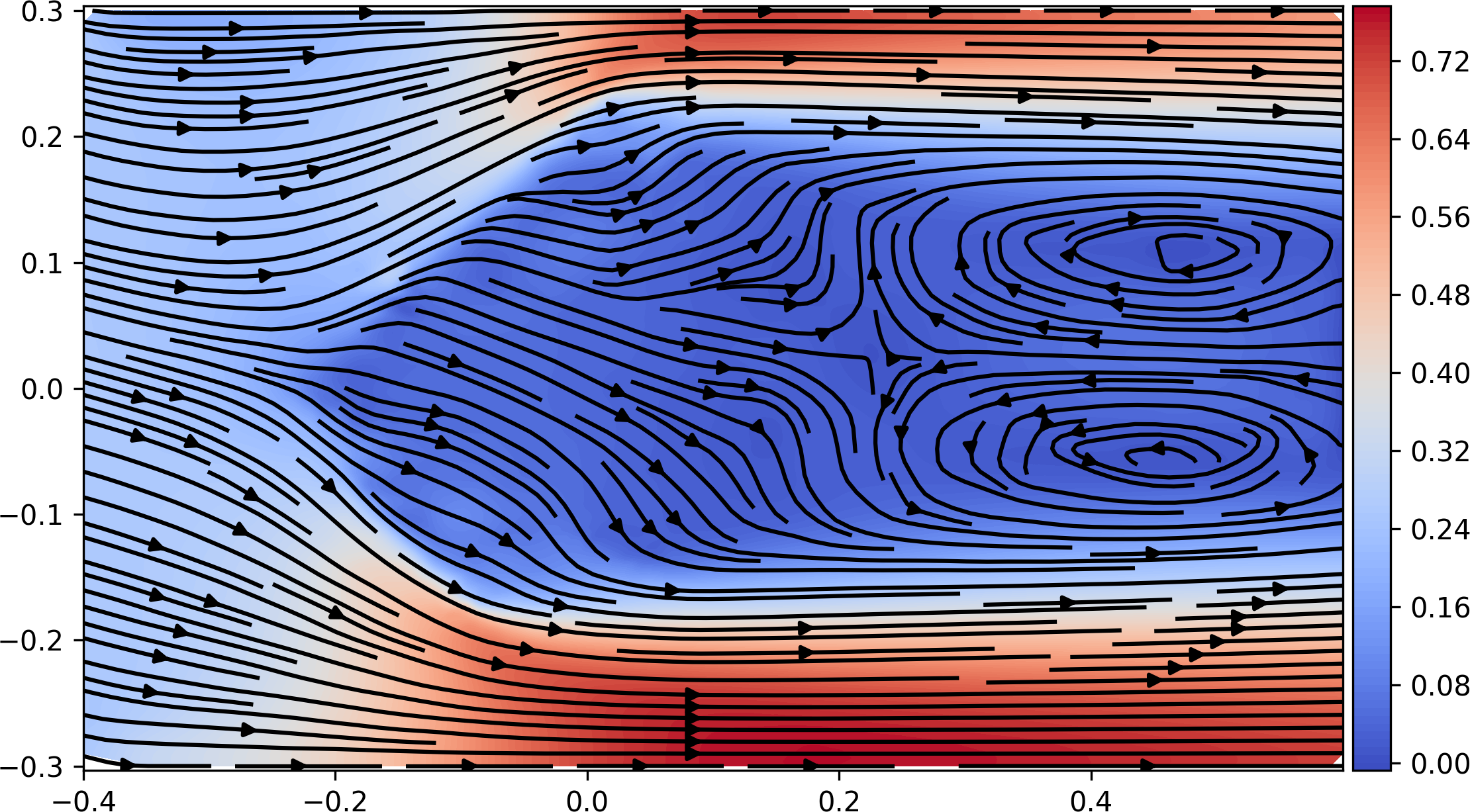}
    \caption{True $U$ }
  \end{subfigure}
  \begin{subfigure}[b]{0.32\textwidth}
    \includegraphics[width=0.99\textwidth]{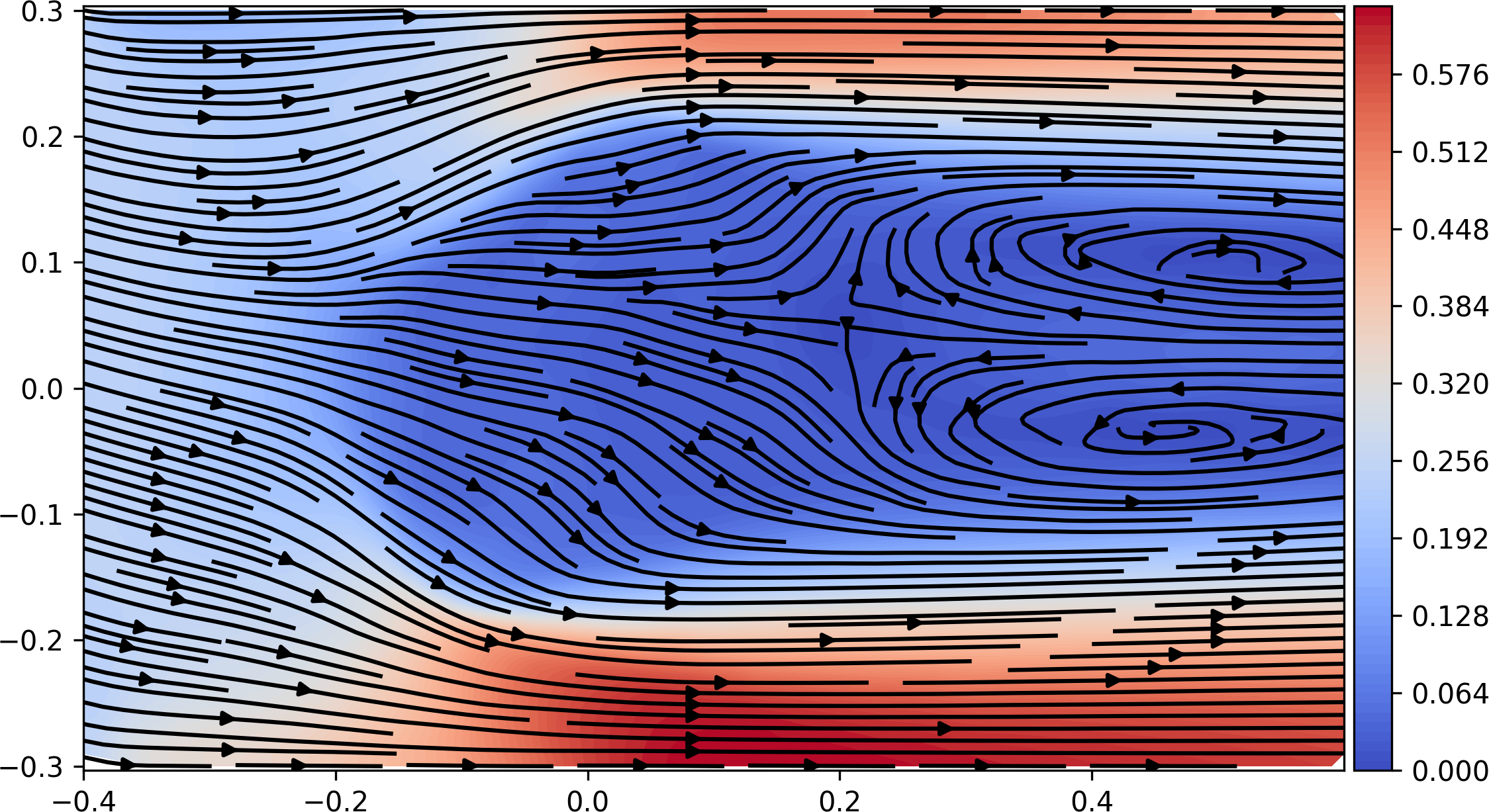}
    \caption{Predicted $U$ }
  \end{subfigure}
  \begin{subfigure}[b]{0.32\textwidth}
    \includegraphics[width=0.99\textwidth]{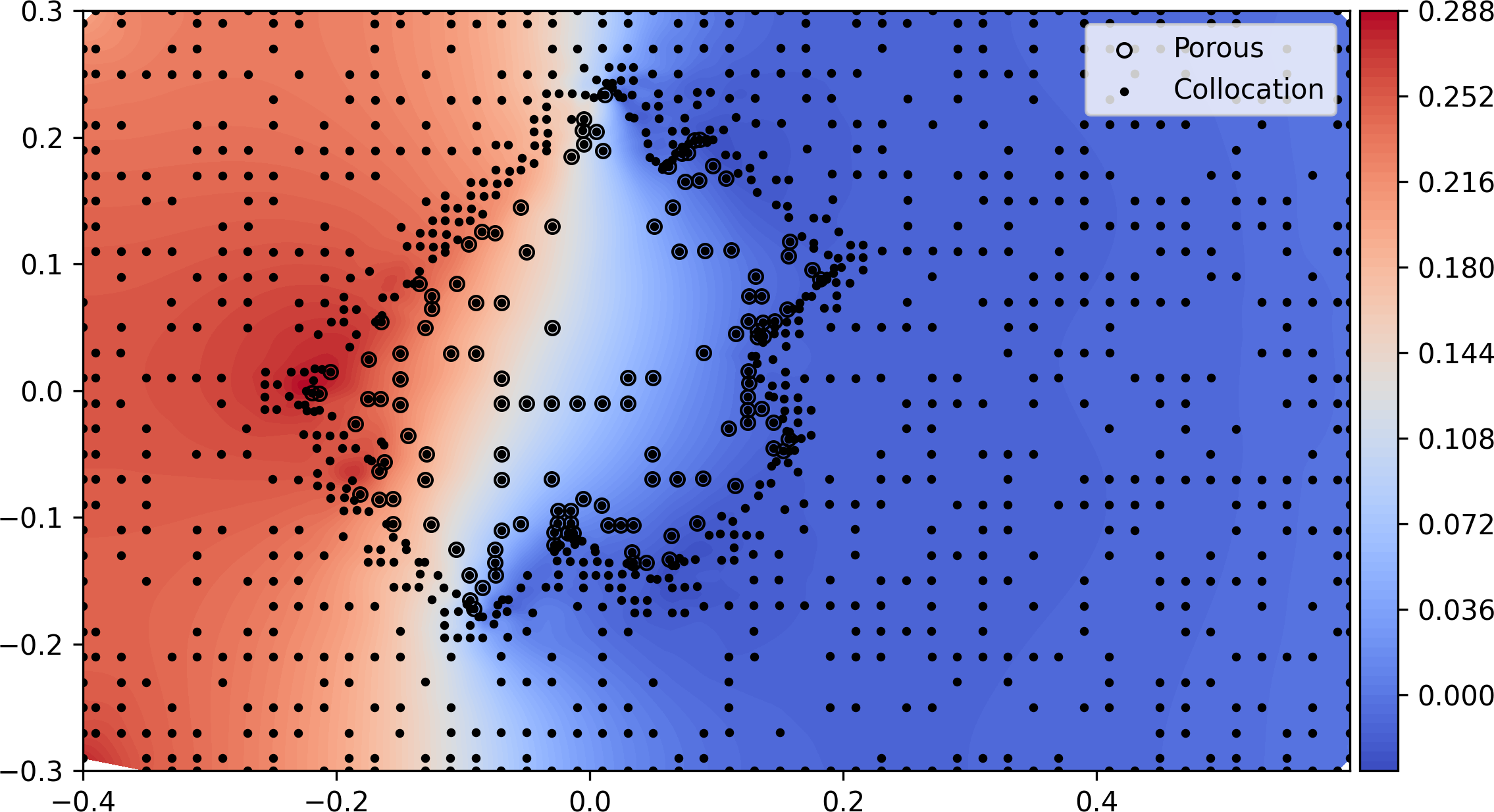}
    \caption{True $p$ }
  \end{subfigure}
  \begin{subfigure}[b]{0.32\textwidth}
    \includegraphics[width=0.99\textwidth]{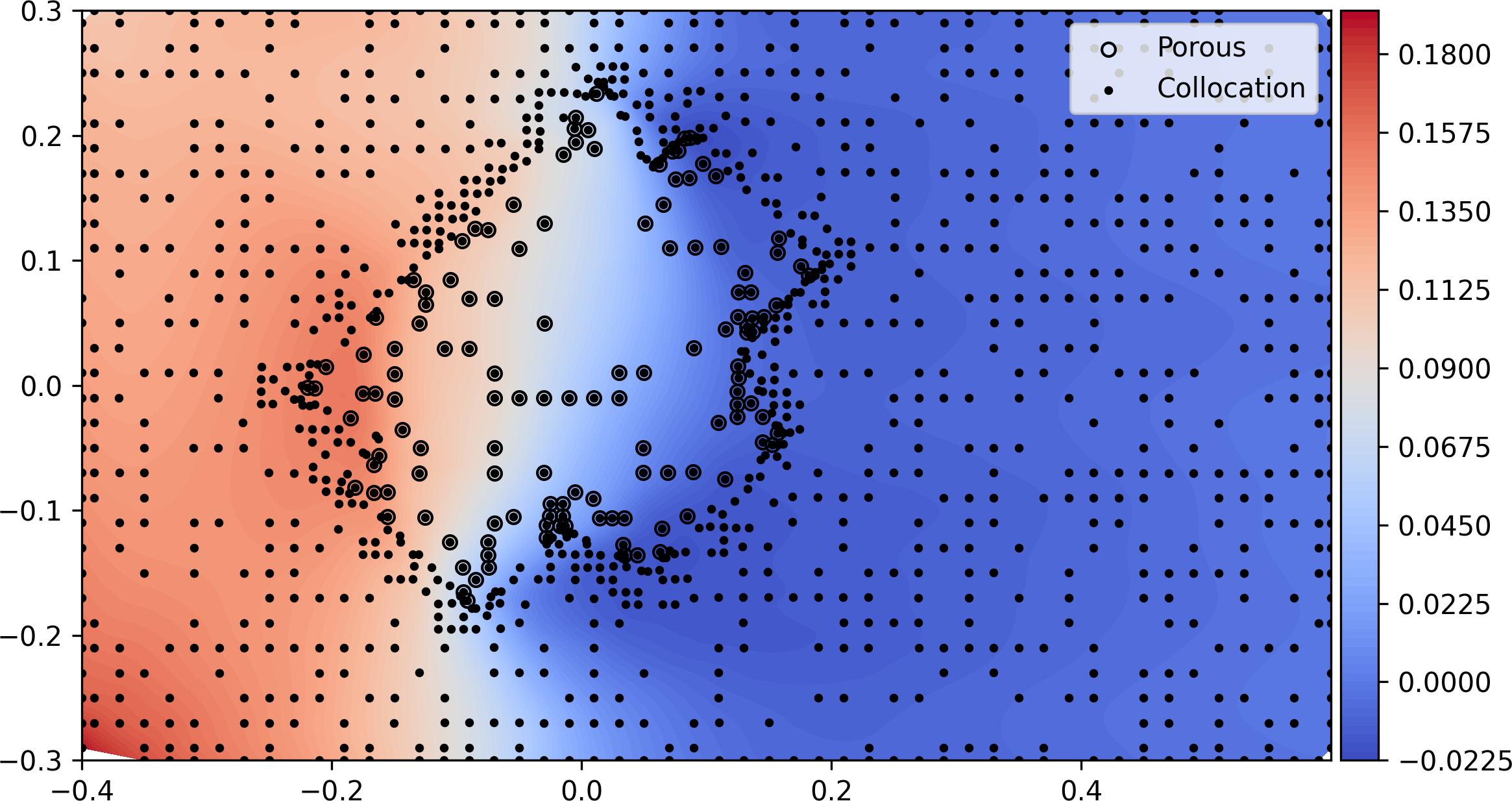}
    \caption{Predicted $p$ }
  \end{subfigure}
  \caption{Errors, prediction and ground truth
    on an unseen geometry
  ($D{=}1.4\, 10^4$, $u_x{=}0.257$ m/s, angle $-15.83^\circ$).}
  \label{fig:pi-gano-plots}
\end{figure}

As expected, errors increase on unseen BCs yet remain acceptable; $u_x$ is again the most sensitive. Fig.~\ref{fig:pi-gano-plots} shows a challenging unseen case (high porosity; obstacle nearly spanning the duct width), with underprediction in high-gradient regions but correct wake reconstruction. Grouping unseen errors by Darcy coefficient (Table~\ref{tab:out-of-range-errors}) shows pressure MAE increasing with $D$, while velocity MAE varies modestly. The average simulation time on the test set is is 0.01s while OpenFOAM took 2.08s.

\begin{table}[tb]
  \centering
  \begin{tabular}{lllll}
    \toprule
    D &  1000 & 2000 & 14000 &16000\\
    \midrule
    $u_x$ & $2.77 \cdot 10^{-2}$ & $1.85 \cdot 10^{-2}$ & $2.36 \cdot 10^{-2}$&$\mathbf{4.16 \cdot 10^{-2}}$\\

    $u_y$ & $6.96 \cdot 10^{-3}$ & $6.94 \cdot 10^{-3}$ & $1.11 \cdot 10^{-2}$&$\mathbf{1.38 \cdot 10^{-2}}$\\

    $p$ & $4.79 \cdot 10^{-3}$ & $1.40 \cdot 10^{-2}$ & $2.11 \cdot 10^{-2}$&$\mathbf{4.49 \cdot 10^{-2}}$\\
    \bottomrule
  \end{tabular}
  \caption{Unseen $D$ average MAE errors.}
  \label{tab:out-of-range-errors}
\end{table}

\subsection{Case 4: 3D porous flow}
We test PI-GANO in 3D (512-dimensional latent geometry space; 16-bit mixed precision). Besides fluid and porous regions, MAE is also reported on solid surfaces to probe boundary fidelity (Table~\ref{tab:windbreak-error}).

\begin{table}[tb]
  \centering
  \begin{tabular}{lllll}
    \toprule
    & Global & Fluid region & Porous region & Solid surface\\
    \midrule
    $u_x$ & $1.15 \cdot 10^{-7}$ &  $1.14 \cdot 10^{-7}$ &  $\mathbf{1.34 \cdot 10^{-7}}$ &  $8.86 \cdot 10^{-8}$ \\
    $u_y$ & $4.14 \cdot 10^{-8}$ &$4.12 \cdot 10^{-8}$ &  $\mathbf{4.44 \cdot 10^{-8}}$ &  $3.89 \cdot 10^{-8}$ \\
    $u_z$ & $3.18 \cdot 10^{-8}$ &$3.10 \cdot 10^{-8}$ &  $\mathbf{4.80 \cdot 10^{-8}}$ &  $2.65 \cdot 10^{-8}$\\
    $p$ & $2.39 \cdot 10^{-12}$ &  $2.15 \cdot 10^{-12}$ & $\mathbf{6.96 \cdot 10^{-12}}$ &  $3.52 \cdot 10^{-12}$ \\
    \bottomrule
  \end{tabular}
  \caption{MAE of the PI-GANO predictions on the windbreaks test set.}
  \label{tab:windbreak-error}
\end{table}

\begin{figure}[tb]
  \centering
  \begin{minipage}{\textwidth}
    \centering
    \begin{subfigure}[b]{0.32\textwidth}
      \includegraphics[width=0.99\textwidth]{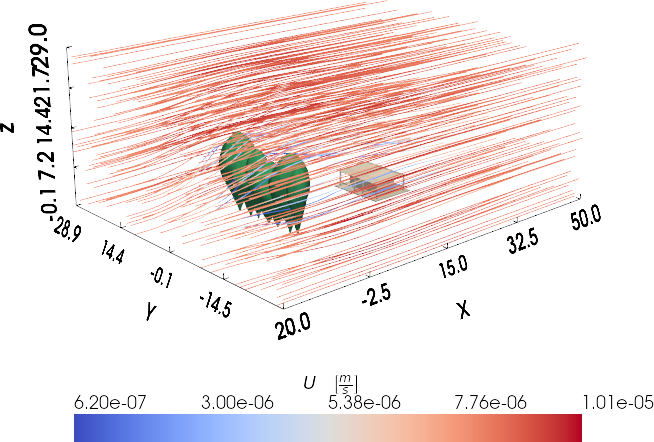}
      \caption{}
    \end{subfigure}
    \begin{subfigure}[b]{0.36\textwidth}
      \includegraphics[width=0.99\textwidth]{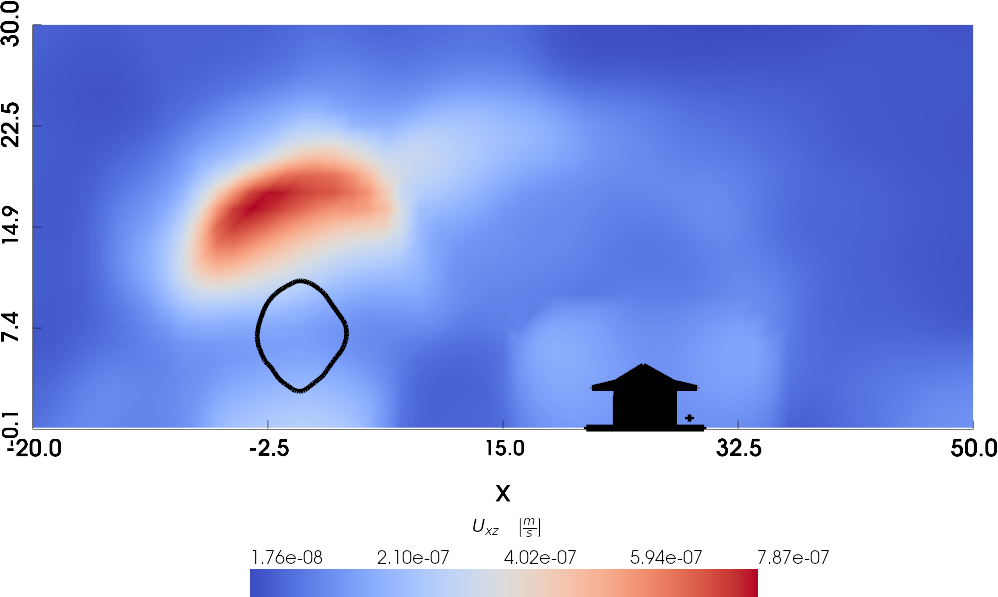}
      \caption{}
    \end{subfigure}
    \begin{subfigure}[b]{0.225\textwidth}
      \includegraphics[width=0.99\textwidth]{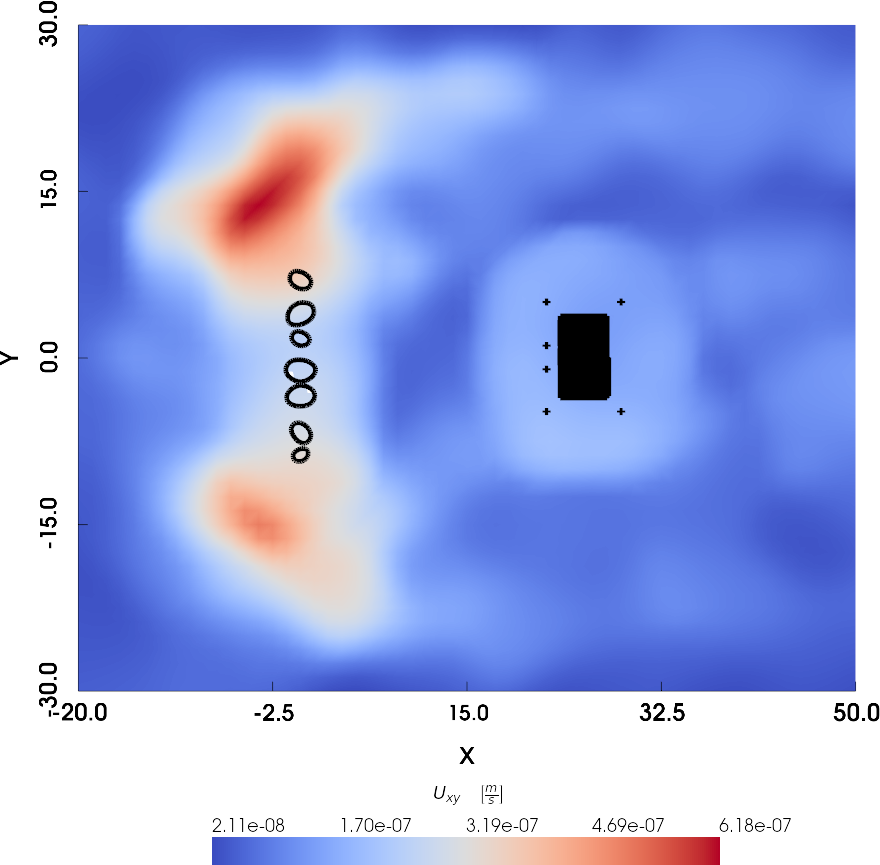}
      \caption{}
    \end{subfigure}
    \begin{subfigure}[b]{0.45\textwidth}
      \includegraphics[width=0.99\textwidth]{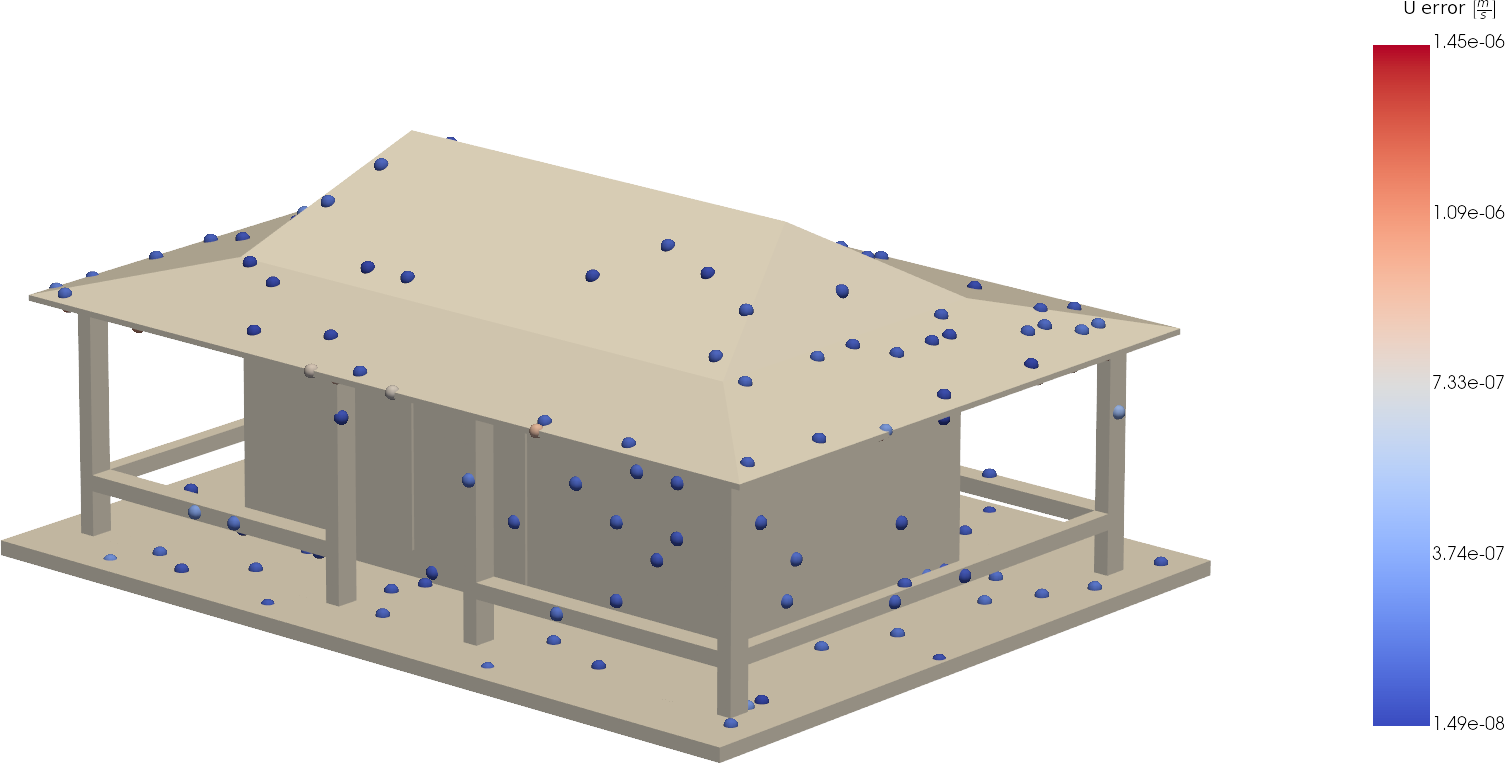}
      \caption{}
    \end{subfigure}
    \begin{subfigure}[b]{0.45\textwidth}
      \includegraphics[width=0.99\textwidth]{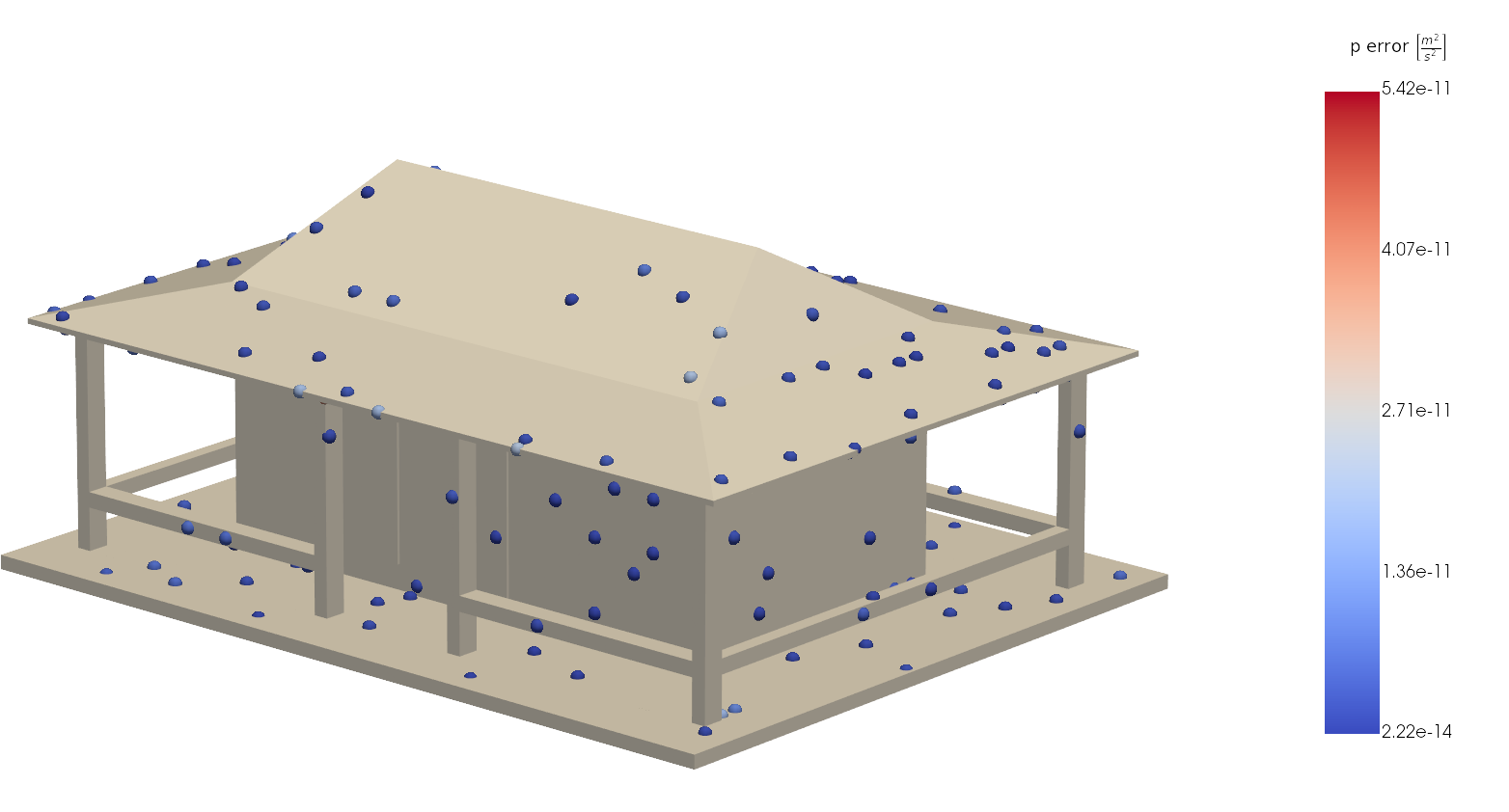}
      \caption{}
    \end{subfigure}
  \end{minipage}

  \caption{Predicted $U$ streamlines (a), sliced xz/xy $U$ errors (b,c), and surface errors on the house for $U$ and $p$ (d,e).}
  \label{fig:windbreak-plots}
\end{figure}

Largest errors occur for $u_x$ and inside tree canopies (complex geometry and high gradients). Still, accuracy is satisfactory given the geometric complexity. Fig.~\ref{fig:windbreak-plots} shows streamlines and surface errors for the worst $u_x$ case (eucalyptus canopy; tapered shape yields stronger impact on the house and complex mixing). The model captures wakes and recirculations; errors concentrate where low-speed wakes interact with free stream. The average simulation time on the test set is is 0.014s while OpenFOAM took 20.04s per case.

\section{Discussion and Conclusions}\label{sec:concolusion}

The proposed methodology allows to effectively model the behavior of mixed fluid-porous medium using PINNs. The tests confirm the ability of the models to generalize not only with respect to variable domain geometries, but also to boundary conditions and material porosity levels.

The experiments prove the effectiveness of the approach in both 2D and more complex 3D cases. Moreover the models were able to predict the flow over a mix of solid and porous materials. Predictions on unseen geometries and boundary conditions show acceptable errors even on cases with configurations not found in the training set.

However, the models proved to have difficulties in obtaining accurate results on flow fields with steep gradients, especially on $u_x$. Several approaches are viable to solve this problem. The first one is the replacement of the geometry encoding block with a PointNet++ based one. This could improve the latent representation of the domain by exploiting the hierarchical architecture proposed in \cite{qi2017pointnet++}. Another strategy could be to use an architecture based on Fourier Neural Operators \cite{li2023geometry,white2023physics}. However judging by the experiment carried out in \cite{li2023geometry}, the model could potentially need a larger number of parameters with respect to the PointNet-based implementations with a consequent increase in training times.

To lift the constraint of low velocities and laminar flow assumptions, it would be necessary to move to a model which takes into account the turbulence of the flow, such as the Reynolds Averaged Navier Stokes equations together with a turbulence closure model such as the $k-\epsilon$ model \cite{LAUNDER1974269}. This would however require to add new terms to the loss, therefore making the optimization problem more difficult, especially with respect to the setting of the scaling parameters. Even if automatic methods have been proposed \cite{liu2024config,BISCHOF2025117914} to balance the loss terms, they are not guaranteed to work on each specific case.

Finally, a sampling strategy could be implemented that takes into account the problematic areas of each experiment, potentially leading to an increase in precision inside those regions.


\section*{Acknowledgments}
Computational resources provided by \href{https://www.disco.unimib.it}{hpc-ReGAInS@DISCo}.
The work was also partially supported the European Union’s Horizon 2020 Research and Innovation Programme under the CISC project (Marie Skłodowska-Curie grant agreement no. 955901);  by the MUSA-Multilayered Urban Sustainability Action project, funded by the European Union-NextGenerationEU, (CUP G43C22001370007, Code ECS00000037); by Project ReGAInS (code 2023-NAZ-0207/DIP-ECC-DISCO23), funded by the Italian University and Research Ministry, within the Excellence Departments program 2023-2027 (law 232/2016); and by the project FAIR - Future AI Research-Spoke 4-PE00000013 D53C22002380006, funded by the European Union-Next Generation EU within the project NRPP M4C2, Investment 1.,3 DD. 341.


\bibliographystyle{unsrt}
\bibliography{refs}

\end{document}